\documentclass[12pt]{article}

\usepackage{graphicx,psfrag,epsf}
\usepackage{enumerate}
\usepackage{natbib}
\usepackage{mathtools}
\usepackage{booktabs}
\usepackage{color}
\usepackage[english]{babel} 
\usepackage[protrusion=true,expansion=true]{microtype} 
\usepackage{amsmath,amsfonts,amsthm}
\usepackage{dsfont}
\usepackage{amssymb}
\usepackage{chngcntr}
\usepackage[toc,page]{appendix}
\usepackage{textcomp}
\usepackage{url}

\usepackage{array}
\usepackage{booktabs} 
\usepackage{natbib}
\usepackage{setspace}

\usepackage{soul}
\usepackage{multirow}
\usepackage{enumitem}
\usepackage{microtype} 
\usepackage[font = small,labelfont=bf,textfont=it]{subcaption}
\usepackage{xcolor}
\usepackage{tabularx}
\newcolumntype{Y}{>{\centering\arraybackslash}X}
\usepackage[font = small,labelfont=bf,textfont=it]{caption} 
\usepackage{footnote}
\usepackage{algorithm}
\usepackage[noend]{algpseudocode}
\usepackage{etoolbox}

\algblock{ParFor}{EndParFor}
\algnewcommand\algorithmicparfor{\textbf{for}}
\algnewcommand\algorithmicpardo{\textbf{do\ parallel}}
\algnewcommand\algorithmicendparfor{\textbf{end\ parallel\ for}}
\algrenewtext{ParFor}[1]{\algorithmicparfor\ #1\ \algorithmicpardo}
\algrenewtext{EndParFor}{\algorithmicendparfor}

\makeatletter
\def\BState{\State\hskip-\ALG@thistlm}

\newcommand{\distas}[1]{\mathbin{\overset{#1}{\kern\z@\sim}}}%
\newcommand{\bm}[1]{\mathbf{#1}}
\newcommand{\bb}[1]{\boldsymbol{#1}}
\newsavebox{\mybox}\newsavebox{\mysim}
\newcommand{\distras}[1]{%
  \savebox{\mybox}{\hbox{\kern3pt$\scriptstyle#1$\kern3pt}}%
  \savebox{\mysim}{\hbox{$\sim$}}%
  \mathbin{\overset{#1}{\kern\z@\resizebox{\wd\mybox}{\ht\mysim}{$\sim$}}}%
}
\newtheorem{theorem}{Theorem}
\newtheorem{proposition}[theorem]{Proposition}

\newcommand{\be}{\begin{equation}}
\newcommand{\ee}{\end{equation}}
\newcommand{\bi}{\begin{itemize}}
\newcommand{\ei}{\end{itemize}}
\newcommand{\ben}{\begin{enumerate}}
\newcommand{\een}{\end{enumerate}}

\newcommand{\R}{\mathbb{R}}
\newcommand{\D}{\mathcal{D}}
\newcommand{\E}{\mathcal{E}}

\newcommand{\U}{\bm U}
\newcommand{\uu}{\bm u}
\newcommand{\V}{\bm V}
\newcommand{\vv}{\bm v}
\newcommand{\Z}{\bm Z}
\DeclarePairedDelimiter\set\{\}

\newcolumntype{K}[1]{\geq {\centering\arraybackslash}p{#1}}
\allowdisplaybreaks
\DeclareMathOperator*{\argmin}{\arg\,\min}
\makeatother

\let\oldbibliography\thebibliography
\renewcommand{\thebibliography}[1]{\oldbibliography{#1}
\setlength{\itemsep}{0pt}} 

\newcommand{\blind}{0}

\patchcmd{\footnotemark}{\stepcounter{footnote}}{\refstepcounter{footnote}}{}{}

\usepackage{booktabs}
\newcommand{\ra}[1]{\renewcommand{\arraystretch}{#1}}
\usepackage{makecell}
\usepackage[section]{placeins}

\begin{document}

\def\spacingset#1{\renewcommand{\baselinestretch}%
{#1}\small\normalsize} \spacingset{1}

\if1\blind
{
  \title{\bf Data Twinning}
  \small
  \author{Akhil Vakayil and V. Roshan Joseph}\hspace{.2cm}\\
  \maketitle
} \fi

\if0\blind
{
  \bigskip
  \bigskip
  \bigskip
  \begin{center}
    {\LARGE \bf Data Twinning}
    \vspace{.25cm}\\
    {Akhil Vakayil and V. Roshan Joseph}\vspace{.2cm}\\
    {Stewart School of Industrial and Systems Engineering\\ 
    Georgia Institute of Technology, Atlanta, GA 30332, USA}\vspace{.2cm}\\
\end{center}
  \medskip
} \fi
\bigskip

\vspace{-0.5cm}
\begin{abstract}
\noindent In this work, we develop a  method named \texttt{Twinning}, for partitioning a dataset into statistically similar twin sets. \texttt{Twinning} is based on \texttt{SPlit}, a recently proposed model-independent method for optimally splitting a dataset into training and testing sets. \texttt{Twinning} is orders of magnitude faster than the \texttt{SPlit} algorithm, which makes it applicable to Big Data problems such as data compression. \texttt{Twinning} can also be used for generating multiple splits of a given dataset to aid divide-and-conquer procedures and $k$-fold cross validation.

\end{abstract}

\noindent
{\it Keywords: Data compression, Data splitting, Testing, Training, Validation.}

\spacingset{1.45} 

\section{Introduction} \label{sec:intro}

Often in statistics and machine learning we are required to partition a dataset, e.g., when $(i)$ splitting a dataset for training and testing, $(ii)$ subsampling from Big Data for conducting tractable statistical analysis or to save storage space, $(iii)$ generating multiple splits of a dataset for divide-and-conquer procedures to act upon, and $(iv)$ creating $k$-fold cross validation sets for model tuning and validation. For this purpose, we propose a novel method named \texttt{Twinning} that can be used for partitioning a dataset into statistically similar sets.

\texttt{Twinning} is motivated from the recent work on optimal data splitting for model validation, by \cite{joseph2021split}. For model validation, the common practice is to randomly split the dataset into training and testing sets, e.g., for an 80-20 split, 20\% of the dataset is selected randomly for testing, while the remaining 80\% is used for training the model. It is easy to see that such random splitting can plausibly give rise to pathological splits, wherein the training and testing sets cover roughly disjoint regions of the feature space, thereby resulting in poor testing performance of the model. Clearly there is a need to systematically split data such that the training set provides sufficient coverage of the feature space. \texttt{CADEX} by \cite{kennard1969computer} marks the beginning of such data splitting procedures in the literature. \texttt{CADEX} pushes majority of the extreme points in the feature space into the training set, while in \texttt{DUPLEX}, a procedure later presented by \cite{snee1977validation} as an improvement over \texttt{CADEX}, the extreme points are about equally distributed between the training and testing sets, thereby producing a more robust testing set for validation. \cite{reitermanova2010data} provides a detailed survey of various data splitting procedures that are geared towards this goal.

\cite{joseph2021split} claim that if rows of a given dataset, including the response, are assumed to be independent realizations from a distribution $\mathcal{F}$, then an unbiased estimate of the model’s generalization error is obtained when the testing set itself is a realization of $\mathcal{F}$.  Random data splitting achieves this distributional similarity, but not \texttt{CADEX} and \texttt{DUPLEX}; furthermore, \texttt{CADEX} and \texttt{DUPLEX} do not include the response in their computations.  The \texttt{SPXY} procedure proposed by \cite{galvao2005method} is similar to \texttt{CADEX}, but includes the response, even still, the distributional similarity is lacking. 

\begin{figure}[h]
\begin{center}
\includegraphics[width = 0.45\textwidth]{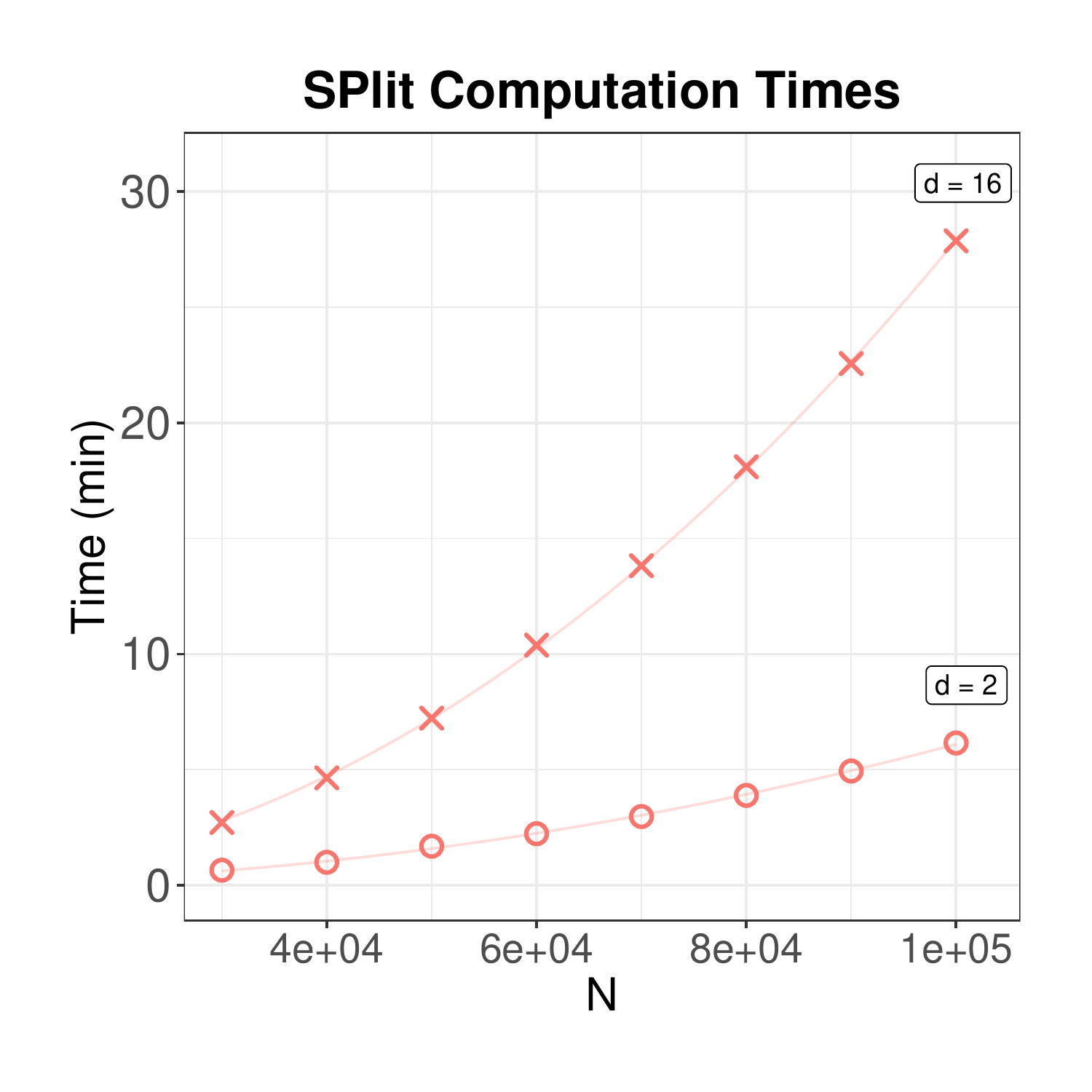}
\caption{Computation times for making an 80-20 split of $d$-dimensional multivariate normal datasets with $N$ rows, using 500 iterations of the original algorithm for \texttt{SPlit}, on a 36-core processor.}
\label{fig:SPlit_complexity}
\end{center}
\end{figure}

The procedure \texttt{SPlit} developed by \cite{joseph2021split} produces testing sets statistically similar to the original dataset. Both parametric and non-parametric models built using \texttt{SPlit} exhibit promising results over random data splitting, i.e., improved and consistent testing performance; only caveat being the computational complexity of making the split. Consider some $d$-dimensional datasets generated by sampling from a multivariate normal with ${\bb \mu} = [0, \dots, 0]^\top \in \R^{d}$ and ${\bb \Sigma} \in \R^{d \times d}: {\bb \Sigma}_{ij} = 0.5^{|i-j|}, \forall i,j \in \{1, \dots, d\}$. Figure \ref{fig:SPlit_complexity} plots the computation time to make an 80-20 split using 500 iterations of the original algorithm for \texttt{SPlit} on a 36-core Intel 3.0 GHz processor, against the size of the dataset. Although the algorithm can be executed in parallel, the quadratic growth suggests a two day wait to split the $16$-dimensional dataset if it had a million rows, even with access to 36 cores. Hence, though apt for model validation, the computational complexity of \texttt{SPlit} holds it back from being applied to Big Data that are prevalent today in every domain. In this work, we develop an efficient algorithm named \texttt{Twinning} that is capable of splitting Big Data with the same objective as \texttt{SPlit}. As will be shown, the computational efficiency of \texttt{Twinning}, coupled with the statistical properties of the splits generated, broadens its applicability to a wide variety of problems that are inaccessible for \texttt{SPlit}.

The remainder of this article is organized as follows. Section \ref{sec:SPlit} reviews the theory behind \texttt{SPlit}. Section \ref{sec:twinning} presents the new \texttt{Twinning} algorithm, and Section \ref{sec:experiments} provides several computational experiments to assess the performance of \texttt{Twinning}. Section \ref{sec:multi-splits} extends \texttt{Twinning} for generating multiple splits of the dataset.  Section \ref{sec:applications} discusses applications of \texttt{Twinning}  in data splitting, data compression, and cross-validation. Finally, we state our concluding remarks in Section \ref{sec:conclusion} .

\section{Review of \texttt{SPlit}} \label{sec:SPlit}
Let $\D = \set{{\bm Z}_i = [{\bm X}_i, Y_i]}_{i=1}^N \in \R^{N \times d}$ be the given dataset, where ${\bm X}_i$ is a $d-1$ dimensional vector representing the $d-1$ features in the $i^{th}$ row, and $Y_i$ denotes the corresponding response value.  Assume that each row of the dataset is independently drawn from a distribution $\mathcal{F}$: 
\[(\bm X_i,Y_i)\overset{iid}{\sim} \mathcal{F},\; i=1,\ldots,N.\]

\noindent The aim is to fit a model $g(\bm X;\bb \theta)$ to the data, where $\bb \theta$ is the unknown set of parameters in the model. To protect against overfitting, we will not use the entire data for estimation. Instead, the dataset is split into two sets: testing set ($\mathcal{D}^1$) and training set ($\mathcal{D}^2$) with sizes $n$ and $N-n$, respectively. Then, $\mathcal{D}^2$ will be used for parameter estimation, and $\mathcal{D}^1$ will be used for assessing the model performance. 

To quantify the model's performance,  define the generalization error as in \citet[Ch.~7]{Hastie2009} by
\begin{equation}\label{eq:err}
    \E=E_{\bm X,Y}\{ L(Y,g(\bm X;\hat{\bb \theta}))|\D^2\},
\end{equation}
where $\hat{\bb \theta}$ is the estimate of $\bb \theta$ obtained from the training set by minimizing a loss $L(Y,g(\bm X;\bb \theta))$. An estimate of the generalization error can be obtained from the testing set as
\begin{equation}\label{eq:errtest}
    \widehat{\E}=\frac{1}{n}\sum_{i=1}^{n}L(Y_i^{1},g(\bm X_i^{1};\hat{\bb \theta})),
\end{equation}
where $\bm U_i=(\bm X_i^1,Y_i^1)$ denote the $i^{th}$ sample in $\D^1$. The estimate of the generalization error will be unbiased if
\begin{equation}\label{eq:condition}
    \bm U_i\sim \mathcal{F},\;i=1,\ldots,n.
\end{equation}

Random sampling is the easiest method to ensure condition (\ref{eq:condition}). However, random sampling gives a high error rate of $\mathcal{O}(n^{-1/2})$ for $|\widehat{\E}-\E|$. \cite{Mak2018} showed that, under some regularity conditions,
\begin{equation}\label{eq:ub}
    |\widehat{\E}-\E|\le C\sqrt{\mathbb{ED}},
\end{equation}
where $C$ is a constant that does not depend on the testing sample, and  $\mathbb{ED}$ is the energy distance \citep{szekely2013energy} between the testing sample and the distribution $\mathcal{F}$. The energy distance can be estimated from the data using
{\small
\begin{align*}
    \overline{\mathbb{ED}}_{n, N} \coloneqq \frac{2}{nN} \sum_{i=1}^n \sum_{j=1}^{N} \|\U_i - \Z_j\|_2 - \frac{1}{n^2} \sum_{i=1}^n \sum_{j=1}^n \|\U_i - \U_j\|_2 - \frac{1}{N^2} \sum_{i=1}^{N} \sum_{j=1}^{N} \|\Z_i - \Z_j\|_2, 
\end{align*}
}%
\noindent where $\|\cdot\|_2$ is the $\ell_2$ norm. Since the energy distance does not depend on the model $g(\cdot;\bb \theta)$ nor the loss function $L(\cdot,\cdot)$, a model-independent testing set can be obtained by minimizing the energy distance. The minimizer of the energy distance is referred to as support points \citep{Mak2018}:
\begin{align}
\set{{\bm z}_i^*}_{i=1}^n &= \argmin_{\set{{\bm z}_i}_{i=1}^n} \ \overline{\mathbb{ED}}_{n, N} \nonumber \\
&= \argmin_{\set{{\bm z}_i}_{i=1}^n} \ \frac{2}{n} \sum_{i=1}^n \mathbb{E}\|\bm{z}_i - \bm{Z}\|_2 - \frac{1}{n^2} \sum_{i=1}^n \sum_{j=1}^n \|\bm{z}_i - \bm{z}_j\|_2.\label{eq:SP}
\end{align}
\cite{Mak2018} also showed that the error rate of $|\widehat{\E}-\E|$ can be reduced to almost $\mathcal{O}(n^{-1})$ by using support points. 

Thus, support points computed from (\ref{eq:SP}) could be used as the testing set, which can work much better than a random sample. This is the idea behind \texttt{SPlit} (Support Points-based split). However, there is one issue. The solution to (\ref{eq:SP}) need not be a subsample of the original dataset, because the optimization is done on a continuous space. What we really need to do is to solve the following discrete optimization problem:
\begin{equation}\label{eq:discSP}
    \set{{\bm U}_i^*}_{i=1}^n=\argmin_{\set{{\bm z}_i}_{i=1}^n\in \D} \ \frac{2}{n} \sum_{i=1}^n \mathbb{E}\|\bm{z}_i - \bm{Z}\|_2 - \frac{1}{n^2} \sum_{i=1}^n \sum_{j=1}^n \|\bm{z}_i - \bm{z}_j\|_2.
\end{equation}
Instead of solving (\ref{eq:discSP}) directly, \cite{joseph2021split} proposed to solve it in two steps: first solve the continuous optimization in (\ref{eq:SP}) using the difference-of-convex (\texttt{DC}) programming technique \citep{Mak2018} to obtain an approximate solution to support points, and then use a sequential nearest neighbor assignment to find the closest points in the dataset to the support points. From here on in, we will refer to this two-step approach as \texttt{DC-NN}, i.e., difference-of-convex programming followed by nearest neighbor assignment. Although much faster than an integer programming solution to (\ref{eq:discSP}), the computational complexity of this approach is still high.

To arrive at the computational complexity of the \texttt{DC-NN} algorithm, we begin with the \texttt{DC} program that has a complexity of $\mathcal{O}(dn(n + N)/P)$ per iteration, where $P$ is the number of processor cores available. Let $\gamma=n/N$. Then, for $\tau$ iterations, we obtain
\begin{align} \label{eq:DC_complexity}
\mathcal{O}(\tau dn(n + N)) &= \mathcal{O}(\tau d\gamma N(\gamma N + N) / P) \nonumber \\
&= \mathcal{O}(\tau d\gamma (\gamma + 1) N^2 / P) \nonumber \\
&= \mathcal{O}(\tau d\gamma N^2 / P) \ \because \gamma \in (0, 1).
\end{align}

\noindent The sequential nearest neighbor assignment can be efficiently performed using a \texttt{kd-tree}. Building the \texttt{kd-tree} is $\mathcal{O}(dN \log{N})$, and $n$ nearest neighbor queries are needed, each with worst case complexity $\mathcal{O}(N^{1 - \frac{1}{d}})$ and average case complexity $\mathcal{O}(\log{N})$. Thus, the average case complexity of the overall procedure can be expressed as

\begin{align}\label{eq:DC-NNcomp}
    \mathcal{O}(dN\log{N} + n\log{N} + \tau d\gamma N^2 / P) &= \mathcal{O}(dN\log{N} + \gamma N\log{N} + \tau d\gamma N^2 / P) \nonumber \\
    &= \mathcal{O}((\gamma + d)N\log{N} +  \tau d\gamma N^2 / P) \nonumber \\
    &= \mathcal{O}(dN\log{N} +  \tau d\gamma N^2 / P) \ \because \gamma < 1  .
\end{align}

The quadratic growth in complexity, with respect to the size of the dataset, is the major computational bottleneck of \texttt{SPlit}. \cite{Mak2018} also provide a version of the \texttt{DC} program that uses stochastic majorization-minimization, where the expectations in $\overline{\mathbb{ED}}_{n, N}$ are computed based on a random sample from $\D$ within each iteration of \texttt{DC}, instead of using the full dataset. Let us denote this procedure as \texttt{SDC}, wherein a random sample of size $\min(\kappa n,N)$ is sampled from $\D$ in every iteration. Thus, for the case of $\gamma<1/\kappa$, the complexity of \texttt{SDC} for $\tau$ iterations reduces to $\mathcal{O}(dn(n + \kappa n) / P) = \mathcal{O}(\tau d\kappa \gamma^2 N^2 / P)$, and the average case complexity of \texttt{SDC-NN} becomes $\mathcal{O}(dN\log{N} +  \tau d \kappa \gamma^2 N^2 / P)$, which is an improvement over \texttt{DC-NN} for small values of $\gamma$. However, for large values of $\gamma$, faster algorithms are needed, which leads us to the main topic of this article that is discussed in the next section.

\section{\texttt{Twinning}} \label{sec:twinning}
The aim of \texttt{Twinning} is to partition a dataset into two  disjoint sets such that they have similar statistical properties. We will call the two sets as \emph{twins}. The twins needn't be of the same size, but they should have similar statistical distributions. Let $\D^1 = \set{\U_i}_{i=1}^n$ and $\D^2 = \set{\V_j}_{j=1}^{N-n}$ be the twins such that $\D^1 \cap \D^2 = \emptyset$ and $\D^1 \cup \D^2 = \D$. Let $\overline{\mathbb{ED}}_{n, N-n}$ be the energy distance between $\D^1$ and $\D^2$, which is given by
{\footnotesize 
\begin{align*}
\overline{\mathbb{ED}}_{n, N-n}  \coloneqq \frac{2}{n(N-n)} \sum_{i=1}^{n} \sum_{j=1}^{N-n} \|\U_i - \V_j\|_2 - \frac{1}{n^2} \sum_{i=1}^{n} \sum_{j=1}^{n} \|\U_i - \U_j\|_2 - \frac{1}{(N-n)^2} \sum_{i=1}^{N-n} \sum_{j=1}^{N-n} \|\V_i - \V_j\|_2.
\end{align*}
}%
\noindent The twins are obtained by minimizing $\overline{\mathbb{ED}}_{n, N-n}$ with respect to $\D^1$ and $\D^2$, i.e.,
\begin{align}
\begin{split}   
\set{{\bm U^*_i}}_{i=1}^n,\set{{\bm V^*_j}}_{j=1}^{N-n}=\argmin_{\set{{\U_i}}_{i=1}^n,\set{{\V_j}}_{j=1}^{N-n}} & \overline{\mathbb{ED}}_{n, N-n} \\
    \text{subject to:} \ & \set{\U_i}_{i=1}^n \cap \set{\V_j}_{j=1}^{N-n} = \emptyset \\
    & \set{\U_i}_{i=1}^n \cup \set{\V_j}_{j=1}^{N-n} = \D. \label{eq:twin}
\end{split}
\end{align}

At first sight, \texttt{Twinning} appears to be a much harder problem than \texttt{SPlit} since we need to perform the optimization in $Nd$ variables instead of $nd$ variables. Interestingly, the following result shows that the objectives of \texttt{Twinning} and \texttt{SPlit} are equivalent.
\begin{proposition} 
Given $n = \gamma N$,
$\overline{\mathbb{ED}}_{n, N} = (1 - \gamma)^2 \cdot \overline{\mathbb{ED}}_{n, N - n}$
\label{prop1}
\end{proposition}

\noindent All the proofs are given in the Appendix. Proposition \ref{prop1} shows that  minimizing the energy distance between $\D^1$ and $\D^2$ is the same as minimizing the energy distance between $\D^1$ and $\D$. Thus, to solve (\ref{eq:twin}), we can solve (\ref{eq:discSP}) and set  $\set{\V^*}_{i=1}^{N-n} = \D \setminus  \set{\U^*}_{i=1}^n$. However, as formally stated below, this is a difficult optimization problem.

\begin{proposition} 
The optimization in (\ref{eq:discSP}) is $\mathcal{NP}$-hard.
\label{prop2}
\end{proposition}

\noindent Hence, from Propositions \ref{prop1} and \ref{prop2}, we have that solving (\ref{eq:twin}) to optimality is intractable. An efficient algorithm to obtain a reasonable solution to (\ref{eq:twin}) is presented next.

\subsection{Algorithm}



There are three parts to $\overline{\mathbb{ED}}_{n, N - n}$ that we will  examine individually, they are
\begin{align}
\begin{split}
    \overline{\mathbb{ED}}_{n, N - n}^1 & \coloneqq \frac{2}{n(N-n)} \sum_{i=1}^n \sum_{j=1}^{N-n} \|\U_i - \V_j\|_2 , \\
    \overline{\mathbb{ED}}_{n, N - n}^2 & \coloneqq - \frac{1}{n^2} \sum_{i=1}^n \sum_{j=1}^n \|\U_i - \U_j\|_2 ,  \text{and} \\ 
    \overline{\mathbb{ED}}_{n, N - n}^3 & \coloneqq - \frac{1}{(N-n)^2} \sum_{i=1}^{N-n} \sum_{j=1}^{N-n} \|\V_i - \V_j\|_2 . \label{eq:energy_split}
\end{split}
\end{align}

\noindent Minimizing $\overline{\mathbb{ED}}_{n, N - n}^1$ translates to bringing $\D^1$ and $\D^2$ closer to each other, whereas minimizing $\overline{\mathbb{ED}}_{n, N - n}^2$ and $\overline{\mathbb{ED}}_{n, N - n}^3$ causes the points in $\D^1$ and $\D^2$, respectively,  to be spread out. Consider the case of \texttt{CADEX} and \texttt{SPXY} that were alluded to in Section \ref{sec:intro}, both procedures effectively aim at minimizing $\overline{\mathbb{ED}}_{n, N - n}^3$ alone, i.e., a well spread out $\D^2$ is produced, while nothing can be said about $\D^1$ and its statistical similarity to $\D^2$. \texttt{DUPLEX} on the other hand focuses on minimizing both $ \overline{\mathbb{ED}}_{n, N - n}^2$ and $ \overline{\mathbb{ED}}_{n, N - n}^3$, still neglecting $ \overline{\mathbb{ED}}_{n, N - n}^1$. \texttt{DUPLEX} is an improvement over \texttt{CADEX} because in essence $\D^1$ is now a relatively rigorous validation set as it is more spread out and potentially includes extreme points. When designing the \texttt{Twinning} algorithm, we will incorporate $\overline{\mathbb{ED}}_{n, N - n}^1$ as well, thereby tying all the three pieces together.

\begin{figure}
\begin{center}
\includegraphics[width = 0.9\textwidth]{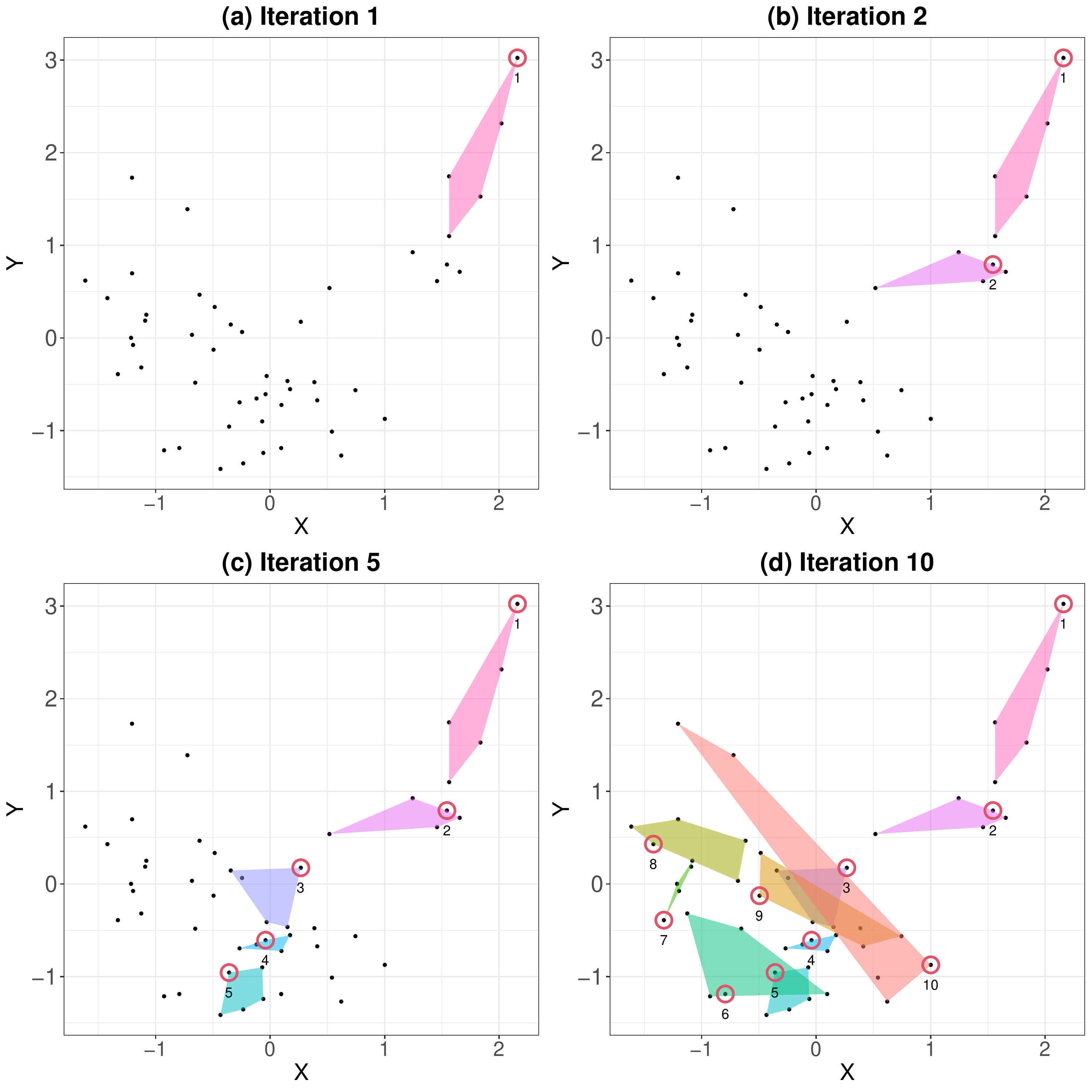} 
\caption{The convex hull of subsets identified by \texttt{Twinning} at the end of iterations 1, 2, 5, and 10 for the sample dataset described in Section \ref{sec:twinning}. Points in $\D^1$ are shown as encircled points, and they are numbered in the order they were selected.}
\label{fig:twinning}
\end{center}
\end{figure}

Assume that $\gamma=1/r$, where $r$ is an integer, i.e., $N = rn$. The dataset $\D$ can now be partitioned into $n$ disjoint subsets each with $r$ points in it. At this stage, if a single point is selected from each of the $n$ subsets into $\D^1$ and the remaining $r-1$ points into $\D^2$, the desired splitting ratio $\gamma$ is achieved. What remains now is to perform the said partitioning, and the selection within the $n$ resulting subsets in a manner that respects the $ \overline{\mathbb{ED}}_{n, N - n}$ criterion.  With this partitioning in mind, we propose a sequential approach where given a starting position $\uu_1 \in \D$, the $r-1$ closest points to $\uu_1$ in $\D$ together with $\uu_1$ form the first of the $n$ subsets. Let this first subset be $\mathcal{S}_1 = \set{\uu_1, \vv_1^{1}, \dots, \vv_1^{r-1}} \in \D$, here $\uu_1$ is pushed into $\D^1$ and the remaining $\vv_1^1, \dots, \vv_1^{r-1}$ are pushed into $\D^2$. For the sake of demonstration, consider a simple two-dimensional dataset with $N=50$ observations generated as follows: $X_i\overset{iid}{\sim} N(0,1)$ and $Y_i|X_i\overset{iid}{\sim} N(X_i^2,1)$ for $i=1,\ldots,N$. We have that $r = 5$ for an 80-20 split. Plot (a) in Figure \ref{fig:twinning} depicts the first subset $\mathcal{S}_1$, given that we start from the encircled point labelled as `1' which represents $\uu_1$.

Next, we select $\uu_2 \in \D \setminus \mathcal{S}_1$ based on some selection rule, and then proceed to identify the $r-1$ closest points to $\uu_2$ in $\D \setminus \mathcal{S}_1$. Let $\mathcal{S}_1 = \set{\uu_1, \vv_1^1, \dots, \vv_1^{r-1}}$ be such that $\|\uu_1 - \vv_1^{k-1} \|_2 \leq \|\uu_1 - \vv_1^{k} \|_2, \forall k = 2, \dots, r-1$, i.e., $\vv_{1}^1$ is the closest neighbor to $\uu_1$ in $\mathcal{S}_1$, while $\vv_1^{r-1}$ is the farthest. Here we propose a selection rule where we select the closest point to $\vv_1^{r-1}$ in $\D \setminus \mathcal{S}_1$ as $\uu_2$. Similar to the first subset, let the second subset be $\mathcal{S}_2 = \set{\uu_2, \vv_2^{1}, \dots, \vv_2^{r-1}} \in \D \setminus \mathcal{S}_1$ where $\uu_2$ is pushed into $\D^1$ and the remaining $\vv_2^1, \dots, \vv_2^{r-1}$ are pushed into $\D^2$. Plot (b) in Figure \ref{fig:twinning} shows both subsets $\mathcal{S}_1$ and $\mathcal{S}_2$ where the encircled point labelled as `2' is $\uu_2$. Now we see the pattern where given $\mathcal{S}_{i-1}$, we select the closest point to $\vv_{i-1}^{r-1}$ in $\D \setminus \cup_{j=1}^{i-1} \mathcal{S}_j$ to be $\uu_i$. It can happen that when $\gamma$ is restricted such that $1/\gamma$ is an integer, $n \neq \gamma N$. In such scenarios, it is straightforward to set $n \leftarrow \lceil \gamma N \rceil $, which results in the cradinality of the final subset to be strictly lower than $r = 1 / \gamma$, i.e., $|\mathcal{S}_n| < r$. Algorithm \ref{alg:algo} formally states the \texttt{Twinning} algorithm. Since the sample dataset has $50$ points, for an 80-20 split, we need to identify 10 subsets: $\mathcal{S}_1, \dots, \mathcal{S}_{10}$  each with 5 points in it as described. Figure \ref{fig:twinning} depicts the subsets identified at the end of iterations 1, 2, 5, and 10 of \texttt{Twinning}. In plot (d) of Figure \ref{fig:twinning}, the encircled points make up $\D^1$, and the remaining points form $\D^2$. 

\texttt{Twinning} attempts to minimize all three parts of $\overline{\mathbb{ED}}_{n, N - n}$ that are outlined in (\ref{eq:energy_split}).  It is immediately seen that, when the $r$ points in each of the $n$ subsets are closer to each other, we end up reducing $\overline{\mathbb{ED}}_{n, N - n}^1$ since for most testing points there are at least $r-1$ of its neighbors in $\D^2$. In other words, for most testing points there will not be another testing point closer than its $(r-1)^{th}$ neighbor, which in turn brings down $\overline{\mathbb{ED}}_{n, N - n}^2$. Reduction in $\overline{\mathbb{ED}}_{n, N - n}^3$ is given if  $\D^2$ is well spread out rather than aggregated in a region. With the proposed selection rule, \texttt{Twinning} sequentially finds subsets $\mathcal{S}_i, \forall i = 1, \dots, n$ such that most subsets are expected to be adjacent to their previous subset with minimal overlaps. This in turn ensures that $\D^2$, which constitutes all but one point from each subset, sufficiently covers the region of the dataset, thereby reducing $\overline{\mathbb{ED}}_{n, N - n}^3$ as alluded. 

\begin{algorithm}
\caption{\texttt{Twinning}}
\label{alg:algo}
\begin{algorithmic}[1]
\State Input $\D \in \R^{N \times d}$, splitting ratio $\gamma : 1/\gamma \in {\mathbb Z}$, and starting position $\uu_1$
\State Standardize the columns of $\D$
\State $n \leftarrow \lceil \gamma N \rceil$; $r \leftarrow 1/\gamma$; $\D^1 \leftarrow \set{}$;  $\D^2 \leftarrow \set{}$
\For{$i \in \set{1, \dots, n}$}
    \If{$i \neq 1$} 
        \State $\uu_i = \argmin_{\uu \in \D} \set{\| \uu - \vv^{r-1}_{i-1} \|_2}$
    \EndIf
    \State \textbf{end if}
    \If{$i = n$} 
        \State $r \leftarrow |\D|$
    \EndIf
    \State \textbf{end if}
    \State $\set{\vv_i^1, \dots, \vv_i^{r-1}} =  \argmin_{\set{\vv^1, \dots, \vv^{r-1}} \in \D \setminus \set{\uu_i}} \set{\sum_{j=1}^{r-1} \|\vv^j - \uu_i \|_2}$
    \State $\D^1 \leftarrow \D^1 \cup \set{\uu_i}$; $\D^2 \leftarrow \D^2 \cup \set{\vv_i^1, \dots, \vv_i^{r-1}}$; $\D \leftarrow \D \setminus \set{\uu_i, \vv_i^1, \dots, \vv_i^{r-1}}$
\EndFor
\State \textbf{end for}
\State \textbf{return} $\D^1, \D^2$
\end{algorithmic}
\end{algorithm}

\begin{figure}
\begin{center}
\includegraphics[width = \textwidth]{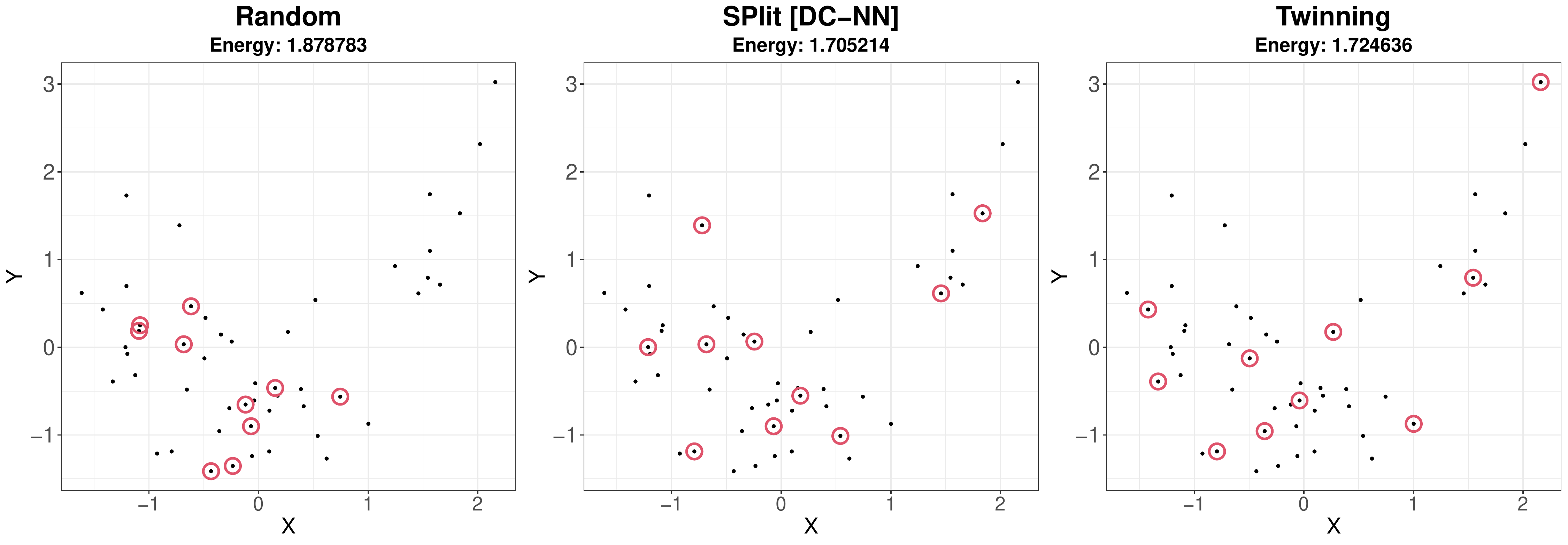} 
\caption{A random split of the dataset described in Section \ref{sec:twinning} is shown on the left, while a \texttt{SPlit} obtained using \texttt{DC-NN} is shown in the middle, and with \texttt{Twinning} on the right. The encircled points constitute $\D^1$ in each of the three plots.}
\label{fig:random_SPlit}
\end{center}
\end{figure}

Figure \ref{fig:random_SPlit} shows a pathological random 80-20 split, a split from 500 iterations of \texttt{DC-NN}, as well as the split obtained by \texttt{Twinning} on the same sample dataset. When comparing different splits of a fixed dataset, it is easier to use $\overline{\mathbb{ED}}_{n, N}$ as the metric, i.e., energy distance between the smaller twin $\D^1$ and the whole dataset $\D$. Similar to how $\overline{\mathbb{ED}}_{n, N-n}$ in (\ref{eq:energy_split}) has three parts, $\overline{\mathbb{ED}}_{n, N}$ also has three parts, of which the third remains constant if the dataset is fixed. Hence, for the remainder of this article, the energy that is being referred to in the plots is
\begin{equation}
    \overline{\mathbb{ED}}_{n, N}^{1, 2} \coloneqq \frac{2}{nN} \sum_{i=1}^n \sum_{j=1}^{N} \|\U_i - \Z_j\|_2 - \frac{1}{n^2} \sum_{i=1}^n \sum_{j=1}^n \|\U_i - \U_j\|_2,
    \label{eq:plot_energy}
\end{equation}

\noindent where $\set{\Z_i}_{i=1}^N$ is the dataset and $\set{\U_i}_{i=1}^n$ is the testing set as before. $\overline{\mathbb{ED}}_{n, N}^{1, 2}$ for the split obtained from \texttt{Twinning} in Figure \ref{fig:twinning} is $1.724636$. We see that the performance of \texttt{Twinning} comes close to that of \texttt{DC-NN}. In Section \ref{sec:experiments}, we provide an extensive comparison between \texttt{Twinning} and \texttt{DC-NN}, where we find that in higher dimensions and for larger values of $N$, \texttt{Twinning} edges out \texttt{DC-NN} with substantially lower computation time.

\subsection{Computational Complexity}
Much of \texttt{Twinning}'s computational complexity can be attributed to nearest neighbor queries. Exact nearest neighbor queries in higher dimensions may not be efficient, and several approximate nearest neighbor search algorithms have been proposed in the literature to address this, e.g., locality-sensitive hashing \citep{slaney2008locality}. \cite{li2019approximate} provides a comprehensive evaluation of nineteen such algorithms.  Nevertheless, following the implementation of \texttt{DC-NN} \citep{SPlitPackage} that uses a \texttt{kd-tree} based exact nearest neighbor search \citep{blanco2014nanoflann}, we use the same in our analysis of \texttt{Twinning}.

Building the \texttt{kd-tree} is $\mathcal{O}(dN \log{N})$. In each of the $n$ iterations, \texttt{Twinning} makes one nearest neighbor query to locate $\uu_i$, $i \neq 1$, and one $r$-nearest neighbors query to identify $\set{\vv^1_i, \dots, \vv^{r-1}_i}$; the former has a worst case complexity of $\mathcal{O}(N^{1 - \frac{1}{d}})$, and $\mathcal{O}(N^{1 - \frac{1}{d}} + r)$ for the latter. The complexity of these nearest neighbor(s) queries is lower in practice, with the average case complexity being $\mathcal{O}(\log{N})$ \citep{friedman1977algorithm}. Updating the \texttt{kd-tree} after each query is an expensive affair, hence, instead of deleting a point after it has been queried out of the tree, the point is merely masked. The overall complexity of \texttt{Twinning} in the worst case can be simplified to 
\begin{align}
    \mathcal{O}(dN\log{N} + 2n(N^{1 - \frac{1}{d}} + r)) &= \mathcal{O}(dN\log{N} + 2\gamma N(N^{1 - \frac{1}{d}} + \frac{1}{\gamma})) \nonumber \\
    &= \mathcal{O}(dN\log{N} + 2(\gamma N^{2 - \frac{1}{d}} +N)) \nonumber \\
    &=  \mathcal{O}(dN\log{N} + \gamma N^{2 - \frac{1}{d}}) , \nonumber
\end{align}

\noindent while the average case complexity is 
\begin{align}
    \mathcal{O}(dN\log{N} + 2n\log{N}) &= \mathcal{O}(dN\log{N} + 2\gamma N\log{N}) \nonumber \\
    &= \mathcal{O}((\gamma + d)N\log{N}) \nonumber \\
    &= \mathcal{O}(dN\log{N}) \ \because \gamma < 1 ,
\end{align}
\noindent which is much better than the complexity of \texttt{DC-NN} given in (\ref{eq:DC-NNcomp}).

\section{Experiments} \label{sec:experiments}
\begin{figure}
\begin{center}
\includegraphics[width = 0.9\textwidth]{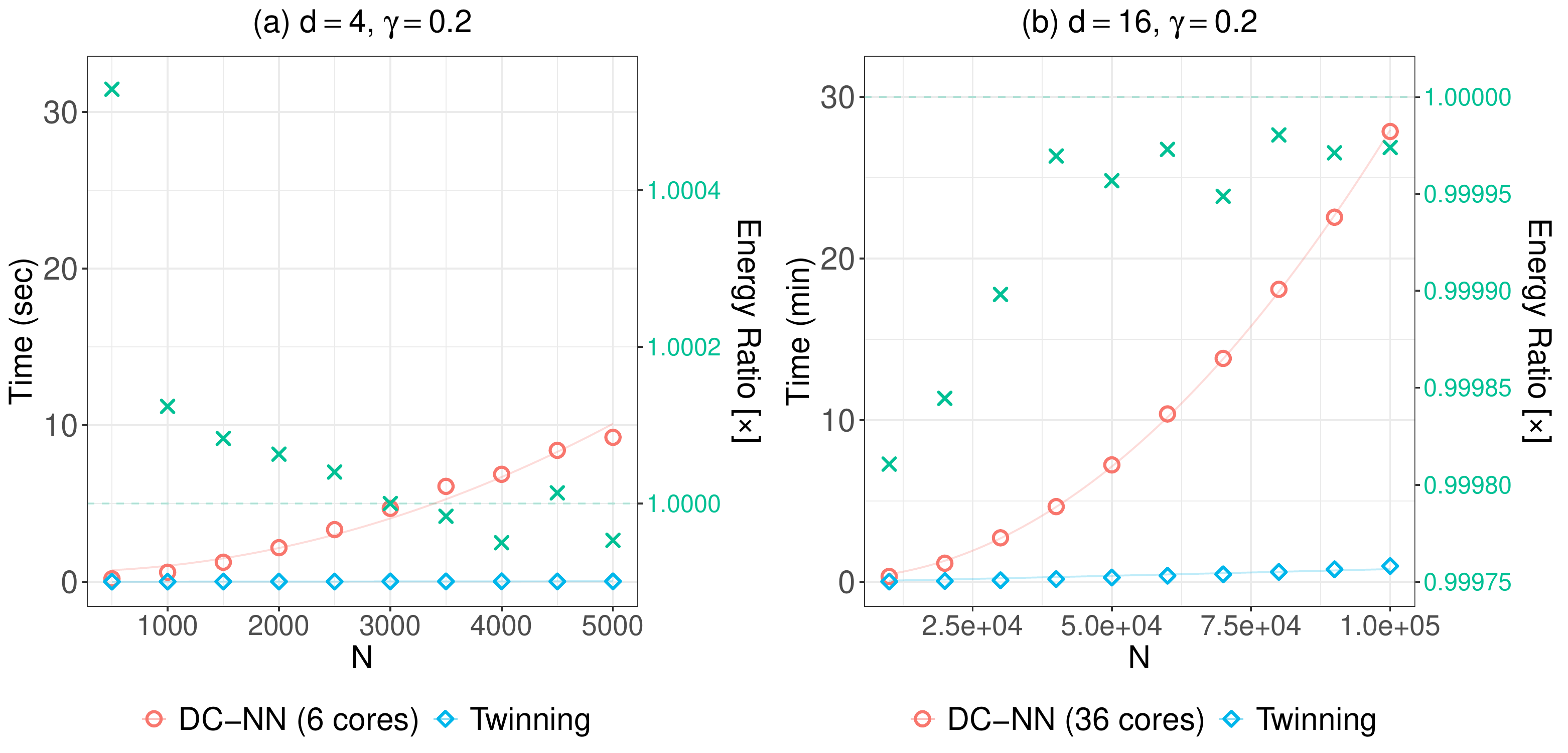} 
\caption{Comparison between \texttt{DC-NN} and \texttt{Twinning} for varying $N$ with fixed $d$ and $\gamma$. 500 iterations are used for \texttt{DC-NN}. The red circles and blue diamonds represent the computation times for \texttt{DC-NN} and \texttt{Twinning}, respectively. The green crosses denote the energy ratio of \texttt{Twinning} to \texttt{DC-NN}.}
\label{fig:exp1}
\end{center}
\end{figure}

\begin{figure}
\begin{center}
\includegraphics[width = 0.9\textwidth]{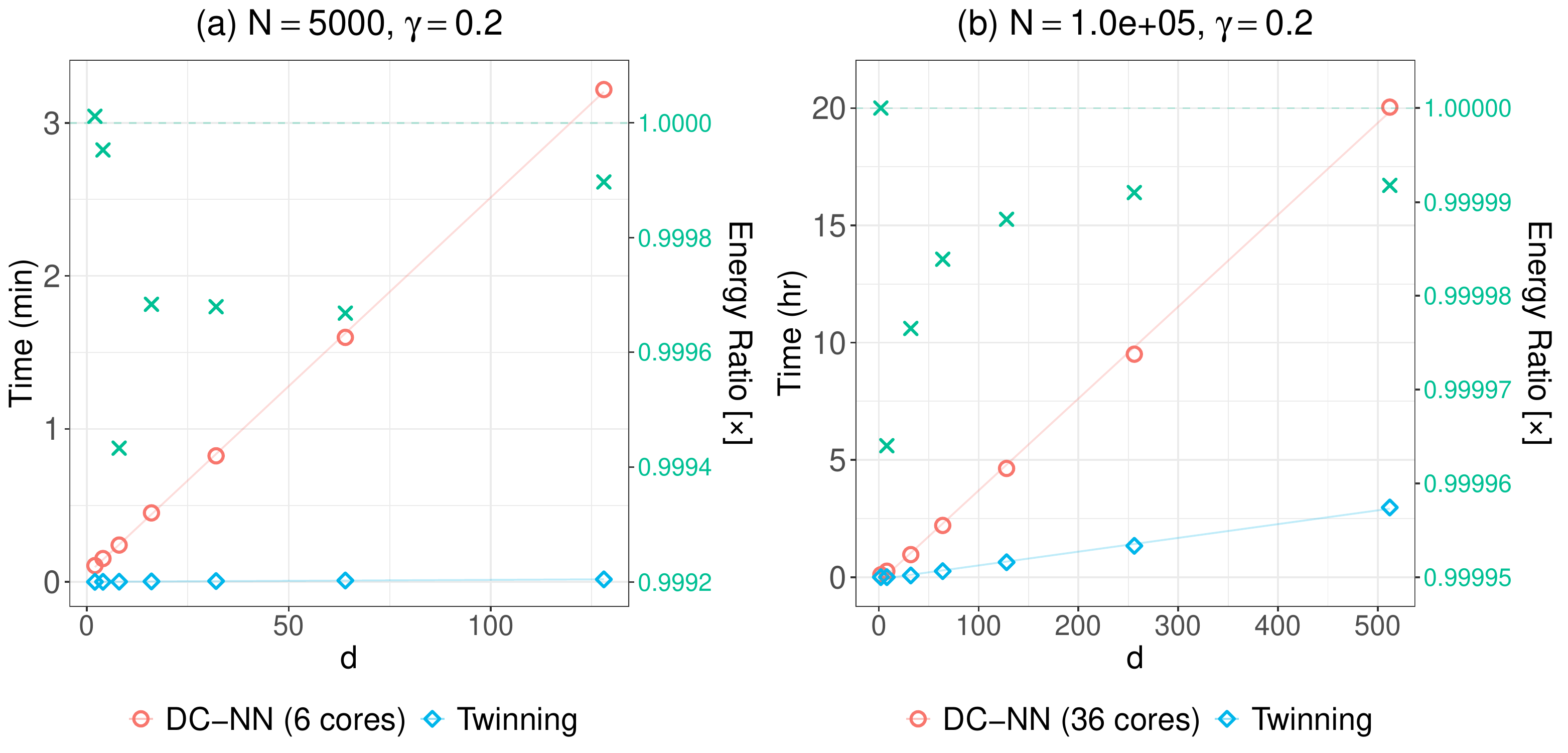} 
\caption{Comparison between \texttt{DC-NN} and \texttt{Twinning} for varying $d$ with fixed $N$ and $\gamma$. 500 iterations are used for \texttt{DC-NN}. The red circles and blue diamonds represent the computation times for \texttt{DC-NN} and \texttt{Twinning}, respectively. The green crosses denote the energy ratio of \texttt{Twinning} to \texttt{DC-NN}.}
\label{fig:exp2}
\end{center}
\end{figure}

\begin{figure}
\begin{center}
\includegraphics[width = 0.9\textwidth]{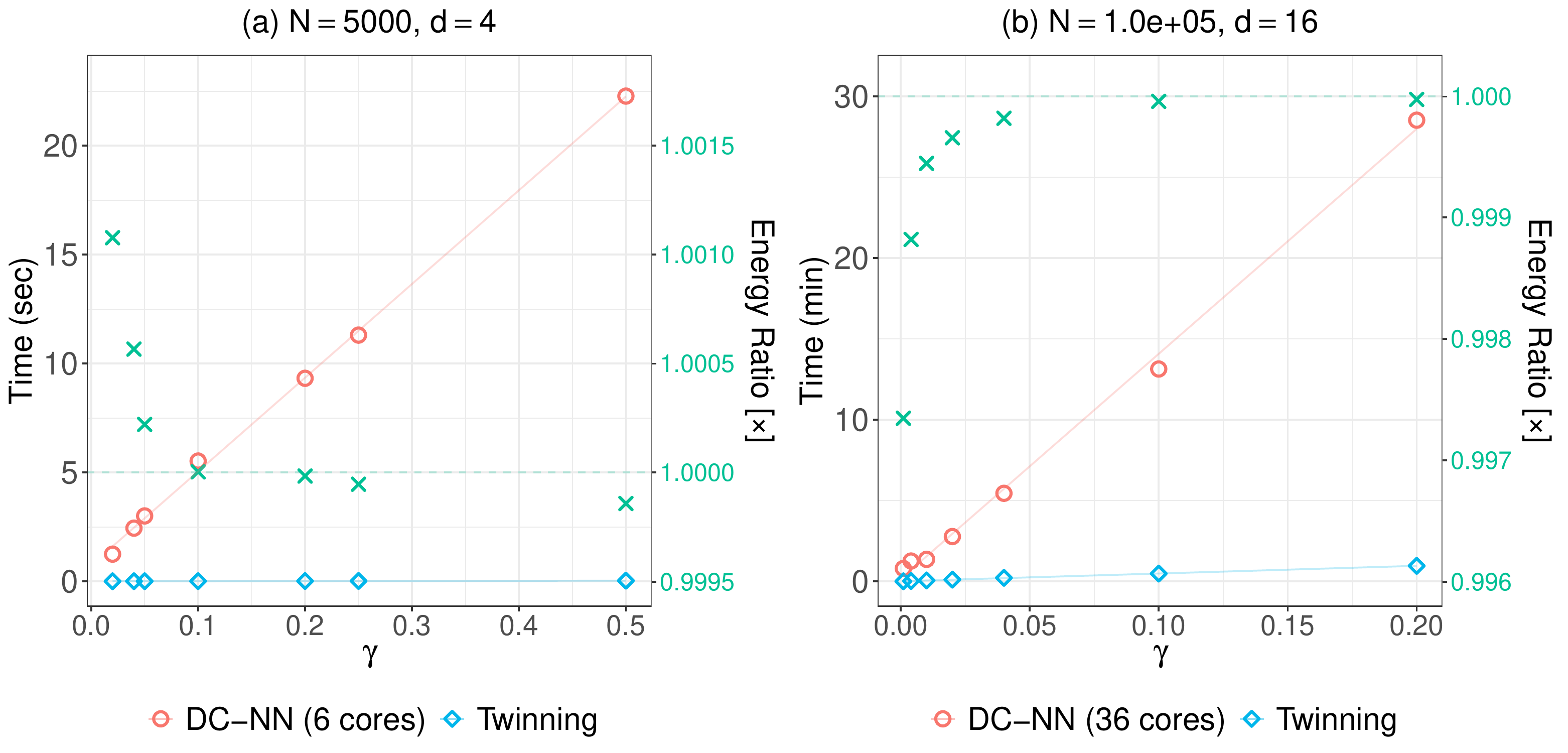} 
\caption{Comparison between \texttt{DC-NN} and \texttt{Twinning} for varying $\gamma$ with fixed $N$ and $d$. 500 iterations are used for \texttt{DC-NN}. The red circles and blue diamonds represent the computation times for \texttt{DC-NN} and \texttt{Twinning}, respectively. The green crosses denote the energy ratio of \texttt{Twinning} to \texttt{DC-NN}.}
\label{fig:exp3}
\end{center}
\end{figure}

Consider $d$-dimensional datasets with $N$ rows generated by sampling from a multivariate normal with ${\bb \mu} = [0, \dots, 0]^\top \in \R^{d}$ and ${\bb \Sigma} \in \R^{d \times d}: {\bb \Sigma}_{ij} = 0.5^{|i-j|}, \forall i,j \in \{1, \dots, d\}$. In this section we compare \texttt{DC-NN} and \texttt{Twinning} for splitting these datasets. The performance of the splits is measured in terms of the computation time and the energy distance metric $\overline{\mathbb{ED}}_{n, N}^{1, 2}$ defined in (\ref{eq:plot_energy}). Plot (a) in Figures \ref{fig:exp1}, \ref{fig:exp2}, and \ref{fig:exp3} considers smaller datasets, where the experiments are run on a laptop with 6-core Intel 2.6 GHz processor, while plot (b) considers bigger datasets, where the experiments are run on a 36-core Intel 3.0 GHz processor. The plots show the ratio of $\overline{\mathbb{ED}}_{n, N}^{1, 2}$ obtained from \texttt{Twinning} to that obtained from \texttt{DC-NN}; hence, an energy ratio of less than 1 for a given $N$, $d$, and $\gamma$ indicates that \texttt{Twinning} performed better than \texttt{DC-NN} with respect to the quality of the split.

\texttt{Twinning} is deterministic when $\uu_1$ is fixed, and in this section $\uu_1$ is chosen to be the point farthest from the centroid of the dataset. With \texttt{DC-NN}, the split can vary over multiple runs of the algorithm owing to its random initialization, hence, for stability, 500 iterations of the iterative algorithm that computes support points is used; the quality of the split improves with every iteration of the iterative algorithm. Moreover, \texttt{DC-NN} can be run in parallel, with the computation time being inversely proportional to the number of parallel threads, whereas \texttt{Twinning} is serial. Still, we see that the computation times for \texttt{Twinning} are significantly lower compared to \texttt{DC-NN}. For example, consider plot (b) of Figure \ref{fig:exp1}, where for $N=10^5$, $d=16$, and $\gamma=0.2$, the computation time for \texttt{Twinning} is better than $\texttt{DC-NN}$ by a factor of 27, even though \texttt{DC-NN} is executed in parallel on 36 cores, i.e., if \texttt{DC-NN} were to be executed on a single core, \texttt{Twinning} would observe an improvement in computation time by a factor of $27 \times 36$ over \texttt{DC-NN}. In addition, the quality of the split obtained by \texttt{Twinning} is better than that obtained from \texttt{DC-NN}.

From Figures \ref{fig:exp1}, \ref{fig:exp2}, and \ref{fig:exp3}, it is evident that \texttt{Twinning} outperforms \texttt{DC-NN}, both in terms of computation time and quality of the split, when it comes to bigger datasets. It is when $N$, $d$, and $\gamma$ are all small, \texttt{DC-NN} produces better quality splits than \texttt{Twinning}. Hence, for partitioning Big Data, \texttt{Twinning} is the algorithm of choice.

\section{Multiplets} \label{sec:multi-splits}
It is often desired to split a dataset into multiple disjoint sets. In keeping with the terminology thus far, if we divide the dataset into three sets with similar statistical properties, we call them \emph{triplets}, four sets as \emph{quadruplets}, and so on. In general, we will call the sets \emph{multiplets}. 

There are several applications for multiplets. The well studied $k$-fold cross validation approach proceeds by randomly splitting a dataset into $k$ sets \citep{Hastie2009}. Moreover, with the technological advances in parallel computing, divide-and-conquer methods that act upon separate blocks of a dataset are becoming increasingly popular for statistical analysis on Big Data \citep{guha2012large}. 

When it comes to $k$-fold cross validation, an estimate of the generalization error of a given model is made based on the individual error estimates on the $k$ sets \citep{rodriguez2010}. The reliability of these $k$ separate error estimates are bound to improve when the $k$ sets are themselves distributed similar to the whole dataset, as shown by \cite{joseph2021split}. The divide-and-conquer methods split Big Data into, say, $k$ manageable sets that are analyzed separately and the results are carefully merged to form a combined inference on the Big Data, thereby circumventing the storage and computational limits of analyzing Big Data on a single machine. It is only reasonable to assume that when these $k$ separate sets are distributed similar to their union, i.e., the Big Data, the quality of the overall inference could improve, albeit further research is needed in this regard. 

Here we demonstrate how \texttt{Twinning} can be adapted to split a given dataset $\D \in \R^{N \times d}$ into $k$ multiplets: $\D^1, \D^2, \dots, \D^k$ such that $\cup_{i=1}^k \D^i = \D$ and $\D^i \cap \D^j = \emptyset, \forall i \neq j$. For ease of presentation, we will explain the methodology to generate multiplets of the same cardinality, i.e., $|\D^i| = N / k, \forall i$. As defined in (\ref{eq:plot_energy}), let $\overline{\mathbb{ED}}_{n_i, N}^{1, 2}$ be the energy of $\D^i$ with respect to $\D, \forall i$, and let $\overline{\mathbb{ED}}_{n, N}^{1, 2 (*)} \coloneqq \max\set{\overline{\mathbb{ED}}_{n_i, N}^{1, 2}: i \in \set{1, \dots, k}}$. A lower $\overline{\mathbb{ED}}_{n, N}^{1, 2 (*)}$ indicates that all the $k$ sets $\D^i, \forall i$ are distributed similar to $\D$. We propose the following three strategies to generate multiplets with \texttt{Twinning},

\begin{itemize}
    \item[$\mathbf{S}^1$] Run \texttt{Twinning} on $\D$ with $\gamma = \frac{n}{N} = \frac{1}{k}$ to obtain $\D^1$ and $\D \setminus \D^1$. Next, run \texttt{Twinning} on $\D \setminus \D^1$ with $\gamma = 1 / (k-1)$ to obtain $\D^2$ and $\D \setminus \cup_{i=1}^2 \D^i$. Repeat until $\D^{k-1}$ and $\D^k$ are obtained. 
    
    \item[$\mathbf{S}^2$] Repeatedly run \texttt{Twinning} on $\D$ with $\gamma = 0.5$ until all the $k$ sets $\D^1, \dots, \D^k$ are obtained, assuming $k$ is a power of 2, i.e., $k = 2^a : a \in \mathbb{Z}$. 
    
    \item[$\mathbf{S}^3$] Run \texttt{Twinning} on $\D$ with $\gamma = \frac{1}{k}$. Let $\mathcal{S}_e, \forall e \in \set{1, \dots, n}$ be the $n$ mutually exclusive and collectively exhaustive subsets of $\D$ identified by \texttt{Twinning}. Let $\mathcal{S}_e = \set{\uu_e, \vv_e^1, \dots, \vv_e^{r-1}}, \forall e$, where $r = k$. For subset $\mathcal{S}_e$, $\forall e$ add $\uu_e$ to $\D^1$, $\vv_e^1$ to $\D^2$, $\dots$, and $\vv_e^{r-1}$ to $\D^k$, $\forall e \in \set{1, \dots, n}$. 
    
\end{itemize}

\begin{figure}[h]
\begin{center}
\includegraphics[width = 0.9\textwidth]{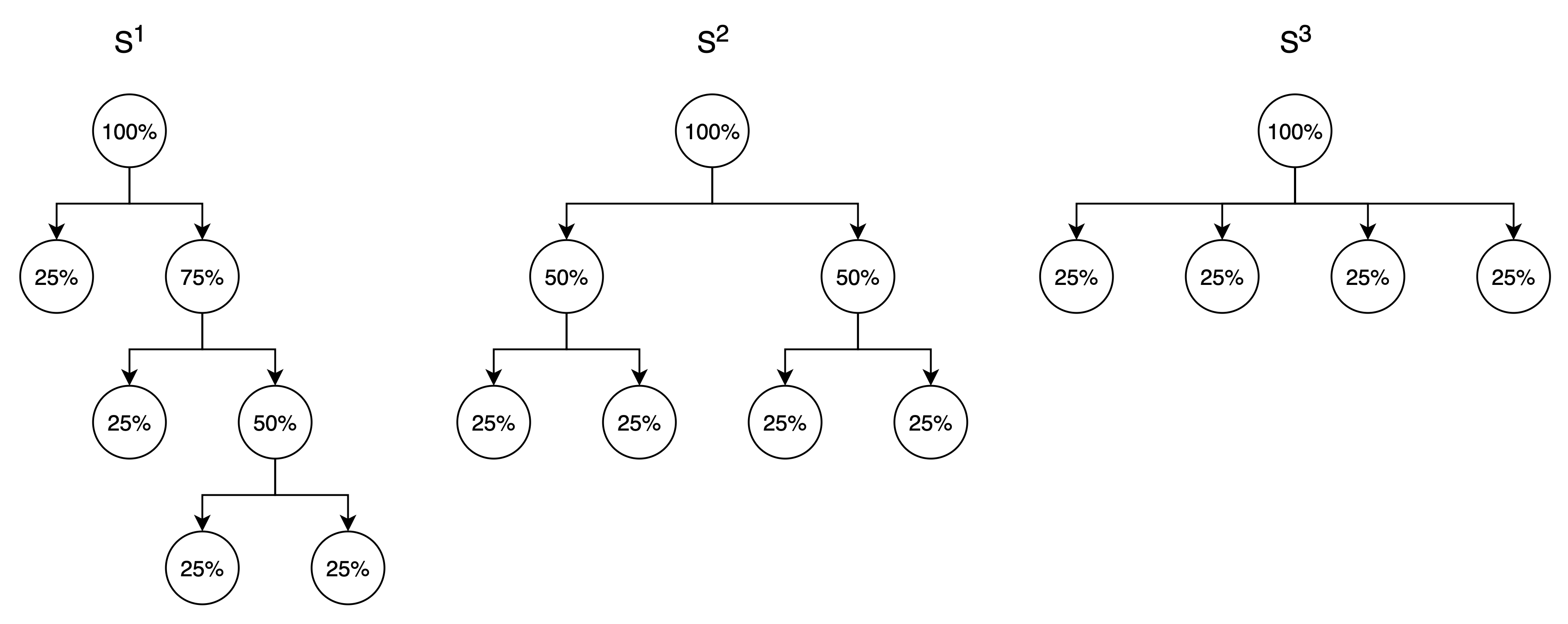} 
\caption{Pictorial representation of the three proposed strategies to generate quadruplets of the same cardinality with \texttt{Twinning}.}
\label{fig:multi_strat}
\end{center}
\end{figure}

\begin{figure}[h]
\begin{center}
\includegraphics[width = 0.45\textwidth]{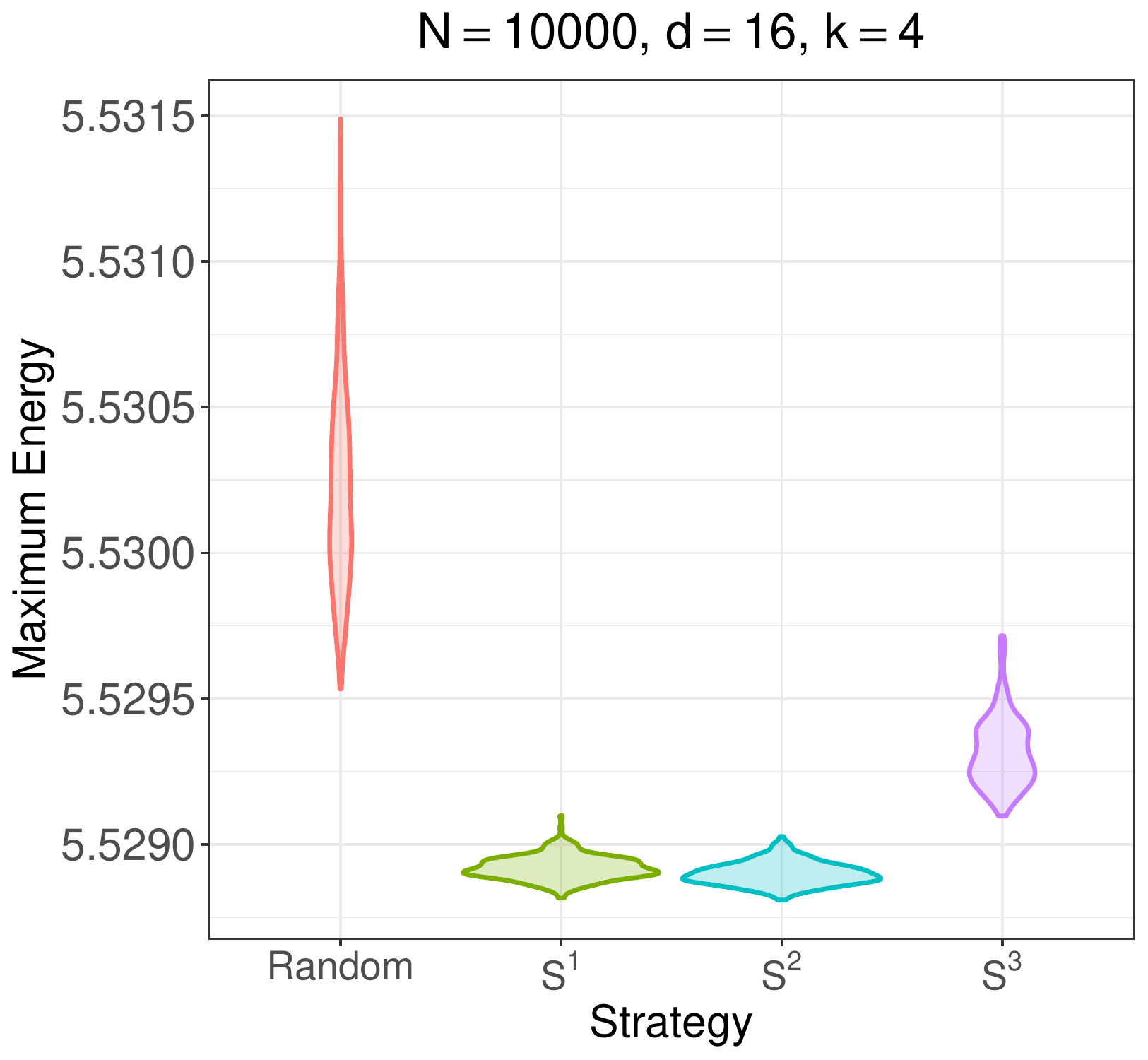}
\caption{Distribution of $\overline{\mathbb{ED}}_{n, N}^{1, 2 (*)}$ over 250 multi-splits of the multivariate normal dataset by random splitting, and the three proposed strategies for generating multiplets with \texttt{Twinning}.}
\label{fig:multi_energy}
\end{center}
\end{figure}

Figure \ref{fig:multi_strat} provides a simple depiction of the three strategies $\mathbf{S}^1$, $\mathbf{S}^2$, and $\mathbf{S}^3$ for generating quadruplets, i.e., $k=4$. We see that both $\mathbf{S}^1$ and $\mathbf{S}^2$ require $k-1$ runs of \texttt{Twinning}, while strategy $\mathbf{S}^3$ gets away with a single run of \texttt{Twinning} on $\D$. To assess the performance of the three strategies, consider a $d = 16$ dimensional dataset with $N = 10000$ rows generated by sampling from a multivariate normal with ${\bb \mu} = [0, \dots, 0]^\top \in \R^{d}$ and ${\bb \Sigma} \in \R^{d \times d}: {\bb \Sigma}_{ij} = 0.5^{|i-j|}, \forall i,j \in \{1, \dots, d\}$. We generate quadruplets of this dataset such that $n_i = 2500, \forall i \in \set{1, 2, 3, 4}$ using $\mathbf{S}^1$, $\mathbf{S}^2$, and $\mathbf{S}^3$, wherein \texttt{Twinning} starts from a random $\uu_1$ in every run. We also make a multi-split randomly for comparison. This experiment is then repeated 250 times on the same dataset, and Figure \ref{fig:multi_energy} reports the distribution  $\overline{\mathbb{ED}}_{n, N}^{1, 2 (*)}$ over these 250 multi-splits with the four strategies that includes random splitting. It is readily observed that all the three strategies $\mathbf{S}^1$, $\mathbf{S}^2$, and $\mathbf{S}^3$ perform better than randomly splitting the dataset into 4 sets such that $\overline{\mathbb{ED}}_{n, N}^{1, 2 (*)}$ is minimized. In addition, we see that strategies $\mathbf{S}^1$ and $\mathbf{S}^2$ perform better than $\mathbf{S}^3$, at the computational cost of additional \texttt{Twinning} runs. However, $\mathbf{S}^3$ can only produce multiplets of same cardinality, whereas it is straightforward to generalize $\mathbf{S}^1$ and $\mathbf{S}^2$ for the case of unequal cardinalities by varying the $\gamma$ in the \texttt{Twinning} runs.

\FloatBarrier
\section{Applications} \label{sec:applications}
In this section, we discuss several applications of \texttt{Twinning}. 

\subsection{Data Splitting} \label{sec:data_splitting}
Consider the problem of predicting taxicab trip duration from the trip characteristics such as total trip distance and proportion of highway. Here, we use the New York City (NYC) taxicab dataset provided as part of the 2017 kaggle competition. In particular, we use the dataset from \cite{taxidata} that contains   $2,074,291$ observations. We will use eight continuous features for modeling, as done in \cite{joseph2021supervised}.

Before fitting the model, we will make an 80-20 split of the dataset, i.e., training set will contain $1,659,432$ observations and testing set $414,859$ observations. If we were to use \texttt{DC-NN}, splitting alone would have taken about one month to finish on our laptop. \texttt{SDC-NN} would also take similar amount of time because it is the same as \texttt{DC-NN} when $\kappa\ge 5$ and $\gamma=0.2$. Thus, it is impractical to optimally split this big dataset using \texttt{DC-NN} or \texttt{SDC-NN}. On the other hand, \texttt{Twinning} was able to make the split in just about two minutes! It is clear that without \texttt{Twinning} one would have to be content with random subsampling. Thus, the remarkable speed of \texttt{Twinning} makes optimal data splitting a reality, especially for Big Data problems.
 
\subsection{Data Compression} \label{sec:data_compression}
In the current data-rich age, computational resources are a bottleneck to analyze or store the vast amount data that is being generated across domains. Furthermore, the trend is set to continue into the foreseeable future, given the lackluster growth rate of computational power over the decade, as we await technological breakthroughs \citep{theis2017end}. Hence, we resort to data compression methodologies that attempt to reduce the size of Big Data while retaining complete or partial information from the Big Data. There exists a fundamental limit to lossless data compression \citep{shannon1948mathematical}, i.e., there is a limit to the extent which a given Big Data can be reduced in size such that no information is lost in the process. The trade-off between the reduction in size and the amount of information retained is what characterizes a lossy data compression that allows for further reduction in size at the expense of information loss. 

\texttt{Twinning} can be viewed as a lossy data compression methodology, where a statistically similar subsample is obtained from the Big Data after compression. The statistical similarity can be associated with retention of information, i.e., the more statistically similar a given subsample is to the Big Data, the more information is retained. Either of the twin subsamples produced by \texttt{Twinning} can be the subsample in question, that can be used instead of the Big Data for tractable statistical analysis. As we will see, the computational efficiency of \texttt{Twinning} is a major factor that enables its use for data compression. 

Since \texttt{Twinning} is based on support points, it can serve as a model-independent data compression methodology, as opposed to the many model-based compression methods in the literature, e.g., information-based optimal subsample (IBOSS) by \cite{Wangetal2019}. Additionally, unlike support points, \texttt{Twinning} produces a subsample of the original dataset, and thus, it could be viewed as a data reduction technique in addition to data compression. We note that, although \texttt{Twinning} makes use of the response column(s) in the dataset, it still behaves like an unsupervised data reduction technique, which is quite different from supervised methods for data reduction, e.g., \texttt{supercompress} by \cite{joseph2021supervised}.

To demonstrate the applicability of \texttt{Twinning} for data compression, consider again the NYC taxi trip dataset introduced in Section \ref{sec:data_splitting}. The training and testing sets have $1,659,432$ and $414,859$ observations, respectively. Suppose we are interested in fitting a random forest regression model using the \texttt{randomForest}  package in \texttt{R} \citep{randomForest}. Fitting a random forest under the default settings of this package runs into memory allocation issues. Therefore, data compression is a must to train the random forest.  Since it is outright impractical to employ \texttt{DC-NN} for the same task, we will compare the performance of \texttt{Twinning} with random subsampling and \texttt{SDC-NN} with $\kappa = 10$. Consider $90\%$ compression, wherein we apply \texttt{Twinning} on the training set with $\gamma = 0.1$. On a laptop with 6-core Intel 2.6 GHz processor, \texttt{Twinning} takes about 40 seconds to reduce the training set, and it  takes 24 minutes to train the random forest on the reduced training set that has $165,944$ observations. However, \texttt{SDC-NN} would have taken more than 11 days for the same task and therefore, it is impractical to use \texttt{SDC-NN} for data compression.

\begin{table}[h]
\centering
\footnotesize
\ra{1.5}
\begin{tabularx}{\textwidth}{Y|YYY|Y|YYY}  
\toprule
\multirow{2}{*}{\textbf{$\gamma$}}  & \multicolumn{3}{c|}{\textbf{Reduction Time }} &  \textbf{Training} & \multicolumn{3}{c}{\makecell[tc]{\textbf{Total Modeling Time}}} \\
& \textbf{Random} & \textbf{\texttt{SDC-NN}} & \textbf{\texttt{Twinning}} &  \textbf{Time} &  \textbf{Random} & \textbf{\texttt{SDC-NN}} & \textbf{\texttt{Twinning}} \\
\midrule
0.1 & 0 & $11^*$ days & 40 sec & 24 min & 24 min & $11^*$ days & $\approx$ 24 min \\
0.05 & 0 &  $2\frac{1}{2}^*$ days &  28 sec & 7 min &  7 min &  $2\frac{1}{2}^*$ days & $\approx$ 7 min \\
0.01 & 0 & 153 min & 15 sec & 27 sec & 27 sec & $\approx$ 153 min & 42 sec \\
0.005 & 0 & 38 min & 14 sec & 11 sec & 11 sec & $\approx$ 38 min & 25 sec \\ 
0.001 & 0 & 101 sec & 13 sec & 2 sec & 2 sec & 103 sec & 15 sec \\
\bottomrule
\end{tabularx}
\caption{Time taken to reduce the NYC taxi trip training set, and then fitting a random forest regression, on a laptop with 6-core Intel 2.6 GHz processor. Fields marked with an $^*$ are estimates.}
\label{tab:compression}
\end{table}

\begin{figure}
\begin{center}
\includegraphics[width = 0.9\textwidth]{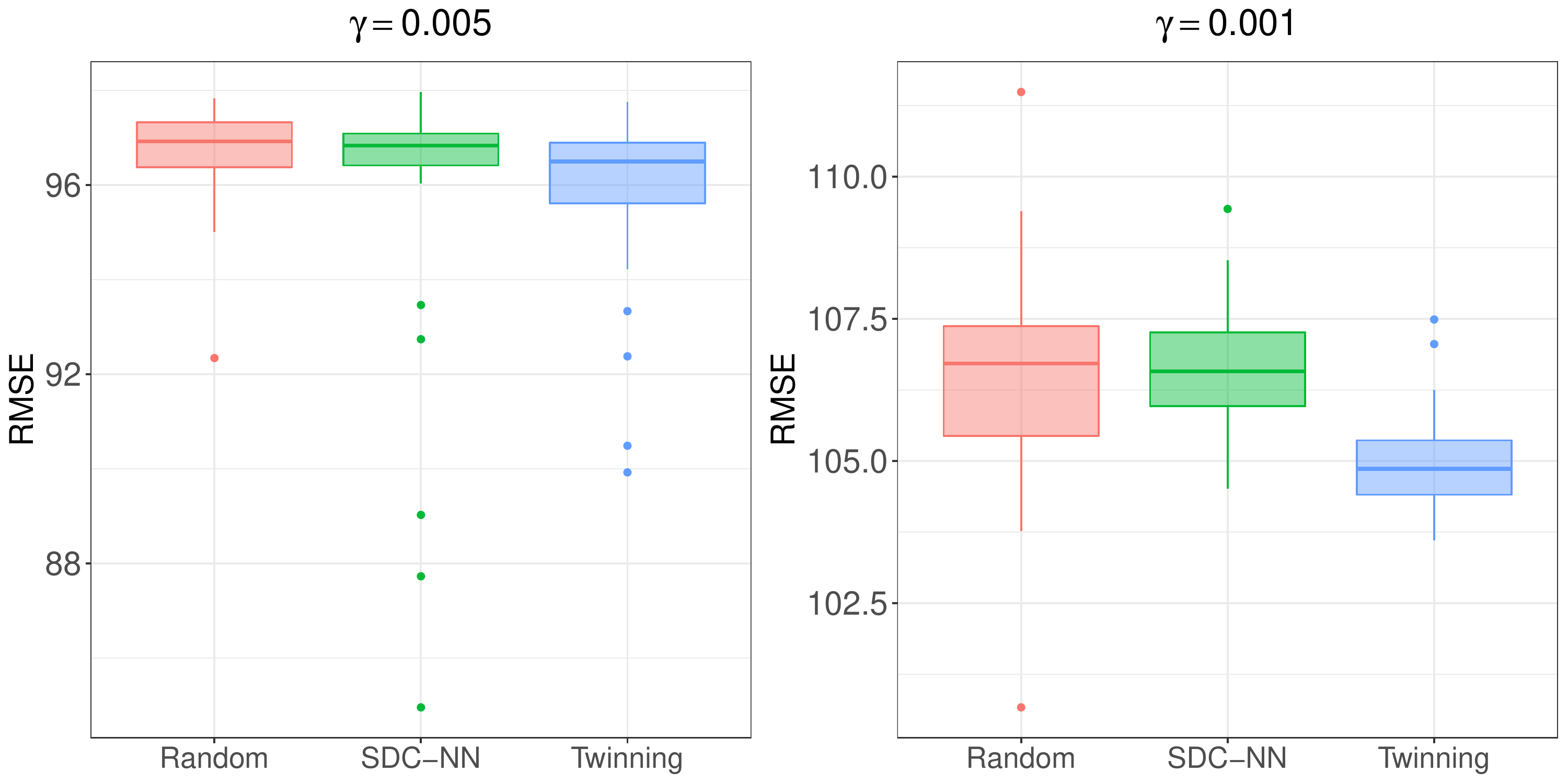} 
\caption{Distribution of testing RMSE over 50 repetitions of compressing the NYC taxi trip training set with random subsampling, \texttt{SDC-NN} ($\kappa = 10$), and \texttt{Twinning}, followed by fitting a random forest regression model.}
\label{fig:nyctaxi}
\end{center}
\end{figure}

Table \ref{tab:compression} lists the total modeling time under the three compression methods: random subsampling, \texttt{SDC-NN} ($\kappa = 10$), and \texttt{Twinning}, for $\gamma$ values: $0.10, 0.05, 0.01, 0.005$, and $0.001$. We can see that the performance of \texttt{Twinning} is quite remarkable--it is almost as fast as the random subsampling-based modeling. On the other hand, \texttt{SDC-NN} is clearly not a feasible data compression method when $\gamma$ is large. To compare the testing performance of the fitted models under the three methods, we repeated the experiment 50 times for $\gamma \in \set{0.005, 0.001}$. The testing performance is measured in terms of the root-mean-squared prediction error (RMSE) on the testing set that has $414,859$ observations. Figure \ref{fig:nyctaxi} provides the distribution of RMSE over the 50 repetitions, where we see that the testing performance of the random forest models fitted with \texttt{Twinning} edges out those with random subsampling and \texttt{SDC-NN}. It is quite amazing to see that \texttt{Twinning} outperforms \texttt{SDC-NN} not only in computational time, but also in the quality of splits.


\subsection{Cross Validation} \label{sec:cross_validation}
In this section, we demonstrate how multiplets can be applied to cross validation, as alluded to in Section \ref{sec:multi-splits}. Consider the airfoil self-noise dataset \citep{brooks1989airfoil} from the UCI Machine Learning Repository. The dataset was originally developed at NASA after a series of aerodynamic and acoustic experiments on airfoil blade sections. The dataset has 1,503 observations with five continuous features and a continuous response indicating the sound pressure level in decibels. We will fit a regression model on the dataset with LASSO \citep{lasso1996}, where the regularization parameter $\lambda$ is estimated by $k$-fold cross validation. 

\begin{figure}
\begin{center}
\includegraphics[width = 0.9\textwidth]{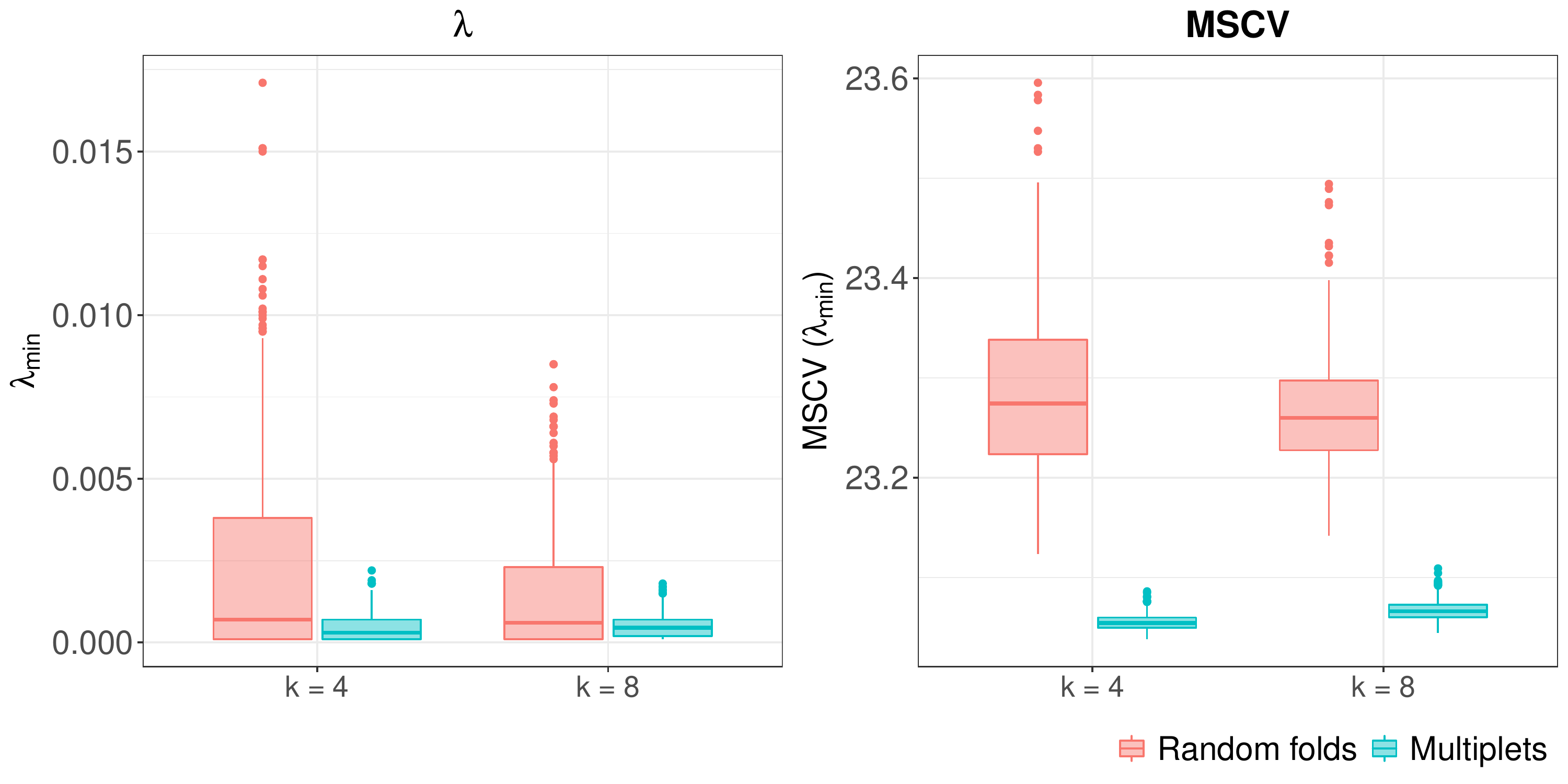} 
\caption{Distribution of $\lambda_{min}$ (left) and MSCV (right) at $\lambda_{min}$, as estimated by LASSO with 4-fold and 8-fold cross validation, over 500 distinct random folds and multiplets of the airfoil self-noise dataset.}
\label{fig:cv}
\end{center}
\end{figure}

Conventionally, the $k$ folds are obtained by randomly partitioning the dataset into $k$ sets of similar size. Here, we analyze the effect of using multiplets, instead of the random partitions as the $k$ folds, e.g., for 4-fold and 8-fold cross validation, we can use quadruplets and octuplets, respectively. The \texttt{glmnet}  \citep{glmnet} package in \texttt{R} is used to fit LASSO, where the same sequence of $\lambda$ values is used when comparing the performance with random folds and multiplets. Let $\lambda_{min}$ be the $\lambda$ corresponding to minimum mean-squared cross validation error (MSCV). Depending upon the $k$ folds supplied to LASSO, the estimated value of $\lambda_{min}$ can vary. Consider fitting LASSO on the given dataset with 4-fold and 8-fold cross validation, the left panel of Figure \ref{fig:cv} depicts the distribution $\lambda_{min}$ estimated by LASSO, over 500 distinct random folds and multiplets, while the right panel shows the distribution of MSCV at $\lambda_{min}$. The multiplets used here are generated by \texttt{Twinning} using the $\mathbf{S}^2$ strategy. 

It is readily observed from Figure \ref{fig:cv} that the use of multiplets for cross validation leads to a stable estimate for the regularization parameter, alongside consistent and smaller MSCV, when compared to random folds. Another interesting takeaway is that even the 4-fold cross validation with multiplets performs better than 8-fold cross validation using random folds, which could prove to be a major computational advantage for tuning computationally expensive models using cross-validation. However, more theoretical investigation and computational experiments are needed to understand this better, which we leave as a topic for future research.

\section{Conclusions} \label{sec:conclusion}
\texttt{Twinning} is built upon the earlier work of \texttt{SPlit} that is aimed at optimally splitting a dataset into training and testing sets. The existing implementation of \texttt{SPlit} uses difference-of-convex programming to find support points of the dataset, and then the testing set is sampled from the dataset using a sequential nearest neighbor algorithm (the overall procedure is termed as \texttt{DC-NN} in this article). This procedure can be time consuming, and thus, not applicable to even moderately large datasets. On the other hand, \texttt{Twinning} directly minimizes the energy distance between the two twin sets, and is several orders of magnitude faster than \texttt{DC-NN}. This computational breakthrough allows \texttt{Twinning} to solve many problems that couldn't even be imagined with \texttt{SPlit}.

We have described multiple potential applications of \texttt{Twinning} besides data splitting, such as data compression and cross-validation, and have demonstrated its performance on a few real datasets. For example, the taxicab dataset containing more than two million data points were split into two sets of 1.66 million and 0.41 million using \texttt{Twinning} in about two minutes on an ordinary desktop computer. This would not have been possible with \texttt{DC-NN}, which would have taken about a month to compute on a similar computer. This computational advantage is crucial for applications such as data splitting and data subsampling. If an optimal procedure for data subsampling takes more time than fitting a computationally expensive statistical model, then it does not make any sense to subsample the data to save model fitting time, one could rather fit the model directly on the full data. The speed at which \texttt{Twinning} can be executed is quite remarkable that it opens the door to a wide variety of problems to which \texttt{Twinning} can be applied, and we expect many more to be discovered in the future.

\appendix
\section*{Appendix}

\subsection*{Proof of Proposition \ref{prop1}}
We have that
{\small
\begin{align}
\overline{\mathbb{ED}}_{n, N} &= \frac{2}{nN} \sum_{i=1}^n \sum_{j=1}^{N} \|\U_i - \Z_j\|_2 - \frac{1}{n^2} \sum_{i=1}^n \sum_{j=1}^n \|\U_i - \U_j\|_2 - \frac{1}{N^2} \sum_{i=1}^{N} \sum_{j=1}^{N} \|\Z_i - \Z_j\|_2 \nonumber \\
\begin{split}
&= \frac{2}{nN} \sum_{i=1}^n \Big\{ \sum_{j=1}^{n} \|\U_i - \U_j\|_2 + \sum_{j=1}^{N-n} \|\U_i - \V_j\|_2  \Big\} - \frac{1}{n^2} \sum_{i=1}^n \sum_{j=1}^n \|\U_i - \U_j\|_2 \\
&\qquad - \frac{1}{N^2} \Big\{ \sum_{i=1}^{n} \sum_{j=1}^{n} \|\U_i - \U_j\|_2 + 2\sum_{i=1}^{n} \sum_{j=1}^{N-n} \|\U_i - \V_j\|_2  + \sum_{i=1}^{N-n} \sum_{j=1}^{N-n} \|\V_i - \V_j\|_2\Big\} \nonumber 
\end{split}\\
&= \frac{2(N-n)}{nN^2} \sum_{i=1}^n \sum_{j=1}^{N-n} \|\U_i - \V_j\|_2 - \frac{(N-n)^2}{n^2N^2} \sum_{i=1}^n \sum_{j=1}^n \|\U_i - \U_j\|_2 - \frac{1}{N^2} \sum_{i=1}^{N-n} \sum_{j=1}^{N-n} \|\V_i - \V_j\|_2 \nonumber \\
&= \frac{(N-n)^2}{N^2} \cdot \Big\{ \frac{2}{n(N-n)} \sum_{i=1}^n \sum_{j=1}^{N-n} \|\U_i - \V_j\|_2 - \frac{1}{n^2} \sum_{i=1}^n \sum_{j=1}^n \|\U_i - \U_j\|_2 \nonumber \\
&\qquad \qquad \qquad \qquad - \frac{1}{(N-n)^2} \sum_{i=1}^{N-n} \sum_{j=1}^{N-n} \|\V_i - \V_j\|_2 \Big\} \nonumber \\
&= \frac{(N-n)^2}{N^2} \cdot \overline{\mathbb{ED}}_{n, N - n} = (1 - \gamma)^2 \cdot \overline{\mathbb{ED}}_{n, N - n} . \nonumber
\end{align}
}%
\begin{flushright} $\qedsymbol$ \end{flushright}

\subsection*{Proof of Proposition \ref{prop2}}
To see that the optimization in (\ref{eq:discSP}) is indeed $\mathcal{NP}$-hard, consider the special case where $n = N/2$, so that we get
{\small
\begin{align}
    \set{{\U}_i^*}_{i=1}^{N/2} &= \argmin_{\set{\U_i}_{i=1}^{N/2} \in \mathcal{D}} \ \frac{4}{N^2} \sum_{i=1}^{N/2} \sum_{j=1}^N \|\U_i - \bm{Z}_j\|_2 - \frac{4}{N^2} \sum_{i=1}^{N/2} \sum_{j=1}^{N/2} \|\U_i - \U_j\|_2 \nonumber \\
    &= \argmin_{\set{\U_i}_{i=1}^{N/2} \in \mathcal{D}} \ \frac{4}{N^2} \sum_{i=1}^{N/2} \Big( \sum_{j=1}^{N/2} \|\U_i - \U_j\|_2 + \sum_{j=1}^{N/2} \|\U_i - \V_j\|_2 \Big) - \frac{4}{N^2} \sum_{i=1}^{N/2} \sum_{j=1}^{N/2} \|\U_i - \U_j\|_2 \nonumber \\
    &= \argmin_{\set{\U_i}_{i=1}^{N/2} \in \mathcal{D}} \ \frac{4}{N^2} \sum_{i=1}^{N/2} \sum_{j=1}^{N/2} \|\U_i - \V_j\|_2, \label{eq:bisection}
\end{align}
}%

\noindent where $\set{\V_j}_{j=1}^{N/2} = \D \setminus \set{\U_i}_{i=1}^{N/2}$. The optimization problem stated in (\ref{eq:bisection}) computes an edge-weighted minimum bisection of a complete graph with $N$ nodes corresponding to the $N$ data points in $\D$, and the  Euclidean distance ($\ell_2$ norm) between the nodes as the edge weights. Since graph bisection is known to be $\mathcal{NP}$-hard for general graphs \citep{garey1974some}, we have that the optimization in (\ref{eq:discSP}) is $\mathcal{NP}$-hard. \begin{flushright} $\qedsymbol$ \end{flushright}

\begin{center}
    {\Large\bf Acknowledgements}

\end{center}

\noindent This research is supported by U.S. National Science Foundation grants DMREF-1921873 and CMMI-1921646.

\bibliography{bibliography}

\begin{thebibliography}{}

\bibitem[Benmeida, 2017]{taxidata}
Benmeida, M. (2017).
\newblock NYC taxi trip durations: Data augmentation using OSRM.

\bibitem[Blanco and Rai, 2014]{blanco2014nanoflann}
Blanco, J.~L. and Rai, P.~K. (2014).
\newblock nanoflann: a {C}++ header-only fork of {FLANN}, a library for nearest
  neighbor ({NN}) with kd-trees.
\newblock \url{https://github.com/jlblancoc/nanoflann}.

\bibitem[Brooks et~al., 1989]{brooks1989airfoil}
Brooks, T.~F., Pope, D.~S., and Marcolini, M.~A. (1989).
\newblock {\em Airfoil self-noise and prediction}, volume 1218.
\newblock National Aeronautics and Space Administration, Office of
  Management~….

\bibitem[Friedman et~al., 2010]{glmnet}
Friedman, J., Hastie, T., and Tibshirani, R. (2010).
\newblock Regularization paths for generalized linear models via coordinate
  descent.
\newblock {\em Journal of Statistical Software}, 33(1):1--22.

\bibitem[Friedman et~al., 1977]{friedman1977algorithm}
Friedman, J.~H., Bentley, J.~L., and Finkel, R.~A. (1977).
\newblock An algorithm for finding best matches in logarithmic expected time.
\newblock {\em ACM Transactions on Mathematical Software (TOMS)},
  3(3):209--226.

\bibitem[Galvao et~al., 2005]{galvao2005method}
Galvao, R. K.~H., Araujo, M. C.~U., Jos{\'e}, G.~E., Pontes, M. J.~C., Silva,
  E.~C., and Saldanha, T. C.~B. (2005).
\newblock A method for calibration and validation subset partitioning.
\newblock {\em Talanta}, 67(4):736--740.

\bibitem[Garey et~al., 1974]{garey1974some}
Garey, M.~R., Johnson, D.~S., and Stockmeyer, L. (1974).
\newblock Some simplified np-complete problems.
\newblock In {\em Proceedings of the sixth annual ACM symposium on Theory of
  computing}, pages 47--63.

\bibitem[Guha et~al., 2012]{guha2012large}
Guha, S., Hafen, R., Rounds, J., Xia, J., Li, J., Xi, B., and Cleveland, W.~S.
  (2012).
\newblock Large complex data: divide and recombine (d\&r) with rhipe.
\newblock {\em Stat}, 1(1):53--67.

\bibitem[Hastie et~al., 2009]{Hastie2009}
Hastie, T., Tibshirani, R., and Friedman, J. (2009).
\newblock {\em The Elements of Statistical Learning: Data Mining, Inference,
  and Prediction}.
\newblock Springer, New York.

\bibitem[Joseph and Mak, 2021]{joseph2021supervised}
Joseph, V.~R. and Mak, S. (2021).
\newblock Supervised compression of big data.
\newblock {\em Statistical Analysis and Data Mining: The ASA Data Science
  Journal}, 14(3):217--229.

\bibitem[Joseph and Vakayil, 2021]{joseph2021split}
Joseph, V.~R. and Vakayil, A. (2021).
\newblock Split: An optimal method for data splitting.
\newblock {\em Technometrics}, 0(0):1--11.

\bibitem[Kennard and Stone, 1969]{kennard1969computer}
Kennard, R.~W. and Stone, L.~A. (1969).
\newblock Computer aided design of experiments.
\newblock {\em Technometrics}, 11(1):137--148.

\bibitem[Li et~al., 2019]{li2019approximate}
Li, W., Zhang, Y., Sun, Y., Wang, W., Li, M., Zhang, W., and Lin, X. (2019).
\newblock Approximate nearest neighbor search on high dimensional
  data—experiments, analyses, and improvement.
\newblock {\em IEEE Transactions on Knowledge and Data Engineering},
  32(8):1475--1488.

\bibitem[Liaw and Wiener, 2002]{randomForest}
Liaw, A. and Wiener, M. (2002).
\newblock Classification and regression by randomforest.
\newblock {\em R News}, 2(3):18--22.

\bibitem[Mak and Joseph, 2018]{Mak2018}
Mak, S. and Joseph, V.~R. (2018).
\newblock Support points.
\newblock {\em The Annals of Statistics}, 46:2562--2592.

\bibitem[Reitermanova, 2010]{reitermanova2010data}
Reitermanova, Z. (2010).
\newblock Data splitting.
\newblock In {\em WDS}, volume~10, pages 31--36.

\bibitem[Rodriguez et~al., 2010]{rodriguez2010}
Rodriguez, J.~D., Perez, A., and Lozano, J.~A. (2010).
\newblock Sensitivity analysis of k-fold cross validation in prediction error
  estimation.
\newblock {\em IEEE Transactions on Pattern Analysis and Machine Intelligence},
  32(3):569--575.

\bibitem[Shannon, 1948]{shannon1948mathematical}
Shannon, C.~E. (1948).
\newblock A mathematical theory of communication.
\newblock {\em The Bell system technical journal}, 27(3):379--423.

\bibitem[Slaney and Casey, 2008]{slaney2008locality}
Slaney, M. and Casey, M. (2008).
\newblock Locality-sensitive hashing for finding nearest neighbors [lecture
  notes].
\newblock {\em IEEE Signal processing magazine}, 25(2):128--131.

\bibitem[Snee, 1977]{snee1977validation}
Snee, R.~D. (1977).
\newblock Validation of regression models: methods and examples.
\newblock {\em Technometrics}, 19(4):415--428.

\bibitem[Sz{\'e}kely and Rizzo, 2013]{szekely2013energy}
Sz{\'e}kely, G.~J. and Rizzo, M.~L. (2013).
\newblock Energy statistics: A class of statistics based on distances.
\newblock {\em Journal of statistical planning and inference},
  143(8):1249--1272.

\bibitem[Theis and Wong, 2017]{theis2017end}
Theis, T.~N. and Wong, H.-S.~P. (2017).
\newblock The end of moore's law: A new beginning for information technology.
\newblock {\em Computing in Science \& Engineering}, 19(2):41--50.

\bibitem[Tibshirani, 1996]{lasso1996}
Tibshirani, R. (1996).
\newblock Regression shrinkage and selection via the lasso.
\newblock {\em Journal of the Royal Statistical Society-Series B}, 58:267--288.

\bibitem[Vakayil et~al., 2021]{SPlitPackage}
Vakayil, A., Joseph, R., and Mak, S. (2021).
\newblock {\em SPlit: Split a Dataset for Training and Testing}.
\newblock R package version 1.0.

\bibitem[Wang et~al., 2019]{Wangetal2019}
Wang, H., Yang, M., and Stufken, J. (2019).
\newblock Information-based optimal subdata selection for big data linear
  regression.
\newblock {\em Journal of the American Statistical Association},
  114(525):393--405.

\end{thebibliography}

\end{document}